\documentclass[preprint,12pt]{elsarticle}

\usepackage{amssymb}
\usepackage{amsmath}
\usepackage{graphicx}   
\usepackage{subcaption} 
\usepackage{booktabs} 
\usepackage{multirow}
\usepackage{makecell}
\usepackage{array}
\usepackage{enumitem}
\usepackage[linesnumbered,ruled,vlined]{algorithm2e}

\journal{Nuclear Physics B}

\begin{document}



\title{MS-MFAD : Multimodal large language models for Face Anti-spoofing Detection} 
\begin{frontmatter}

\author[a,b]{XiaoyongYu}
\ead{xiaoyongyu666@gmail.com} 
\author[a]{RongzhenLi}
\author[a]{ShumingShi}
\author[b]{XingeYou\corref{cor1}} 
\ead{youxg@mail.hust.edu.cn}
\address[a]{Mashang Consumer Finance Co., Ltd., 4th to 8th Floors, Building B2, Yuxing Plaza, No. 52 Huangshan Avenue Middle Section, Chongqing, 401121, Chongqing, China}
\address[b]{Huazhong University of Science and Technology, 1037 Luoyu Road, Hongshan District, Wuhan, 430074, Hubei, China}
\cortext[cor1]{Corresponding author.} 

\begin{abstract}
Facial biometric recognition systems currently face compound threats intertwining generative AI and high-fidelity physical spoofing. Existing defenses suffer from systemic bottlenecks, including poor generalization, non-auditable reasoning, and reliance on massive, low-quality datasets. To address these challenges, we propose Multimodal Large Language Models (MFAD) for face anti-spoofing detection, an explainable reasoning system for Unified Face Anti-Spoofing Detection (UFAD), accompanied by a semantic-level annotation benchmark. Unlike methods relying on external tools or coarse alignment, MFAD activates the intrinsic reasoning capabilities of Multimodal Large Language Models (MLLMs) via a fine-grained pixel-semantic anchoring mechanism. This eliminates localization hallucinations and ensures auditable reasoning paths. We introduce a cross-attack semantic-level unified annotation paradigm: by annotating only 1,000 precise masks per attack category we generate reasoning evidence chains strictly corresponding to spoofed regions. Supervised fine-tuning on the Qwen-VL foundation model demonstrates that, using limited high-quality samples, the system achieves a 40–50\% relative reduction in in-domain ACER and restricts cross-domain performance degradation to within 11.62\%/5.23\%, significantly outperforming existing frameworks. Furthermore, under white-box adversarial attacks, detection accuracy drops by only 3.2\%, validating the robustness of semantic anchoring compared to models trained on massive short-text data. Domain practitioners rated the evidence reliability of reasoning paths at 4.57/5, with inference latency satisfying real-time deployment requirements. These results confirm that a few-shot, high-quality semantic annotation paradigm is effective for building trustworthy, explainable, and cost-efficient UFAD systems.
\end{abstract}



\end{frontmatter}



\section{Introduction}
As the core pillar of digital identity authentication, facial biometric technology has been deeply integrated into critical domains such as financial payments, public security, and smart terminals\cite{kuznetsov2024attacknet}\cite{jingade2023extended}. However, with the rapid proliferation of Generative AI and high-fidelity 3D printing technologies, UFAD is facing unprecedented compound security threats. Attack methods have evolved from early static photographs and video replay attacks\cite{liu2018learning} to hybrid forms encompassing deepfake digital face swapping\cite{rossler2019faceforensics++}\cite{li2020celeb}\cite{dolhansky2019deepfake}, 3D hyper-realistic masks\cite{erdogmus2013spoofing}\cite{liu20213d}, and high-definition screen texture replay\cite{li2019replayed}. The intertwining of physical presentation attacks and digital forgery attacks not only significantly increases the success rate of bypassing detection but also poses severe Visual-Linguistic Face Forgery Detection (VLFFD) challenges to the generalization capabilities and judicial admissibility of existing defense systems. Although early methods based on handcrafted features\cite{maatta2011face}\cite{lin2019face} and traditional deep learning\cite{atoum2017face}\cite{zheng2021attention}\cite{yu2020searching} have achieved certain progress on individual datasets, constructing a UFAD system that can simultaneously defend against multiple attack types, possesses auditable reasoning capabilities, and can be deployed at low cost remains a critical challenge (to be urgently resolved) in the current security landscape.

In recent years, MLLMs have been introduced to the task of face anti-spoofing. Investigating proprietary compact MLLMs models is of considerable necessity in face anti-spoofing\cite{pathak2026hybrid}\cite{sun2022privacy}. Deploying large cloud-hosted models requires transmitting sensitive facial biometric data to remote servers, introducing data leakage risks, while imposing stringent demands on GPU memory and network stability—any service disruption may paralyze the entire authentication pipeline. Moreover, face anti-spoofing demands millisecond-level inference latency for seamless user experience\cite{zhang2024triplet}. In contrast, compact MLLMs models can be deployed directly on edge devices, ensuring facial data never leaves the local domain. By virtue of their reduced parameter count and lower computational complexity, they enable real-time inference on consumer-grade hardware without cloud dependency, constituting a critical pathway for practical face anti-spoofing deployment\cite{yu2025learning}.

However, existing approaches still exhibit significant gaps in security capabilities. Due to the expansive feature space of MLLMs and the significant distribution discrepancy in UFAD training data, MLLMs are inherently well-suited for UFAD tasks, which integrate digital and physical functionalities into a unified model\cite{he2024joint} \cite{liu2024moeit}. Due to the strong reasoning and generalization capabilities of MLLMs, researchers have initially explored zero-shot transfer learning. Uncertainty-guided cross-adapter have improved zero-shot transferability, their alignment granularity remains restricted to the image-level or coarse region-level\cite{lin2025reliable}. Consequently, they fail to provide pixel-level forgery evidence admissible in forensic contexts and are prone to localization hallucinations when confronted with micro-artifacts, such as 3D mask material anomalies or screen moiré patterns. The facial parsing capabilities of FaceXformer, proposed by Narayan et al., are limited to facial regions and fail to capture contextual forgery cues in clothing and backgrounds\cite{narayan2025facexformer}.

Researchers first conducted Continual Pre-Training (CPT) on existing multimodal models using large-scale image-text paired data.  MS-UFAD\cite{jiang2025ms}\footnote{MS-UFAD, where MS denotes Mashang Consumer Finance Co., Ltd., and UFAD refers to Unified Face Anti-spoofing Detection.} and UniAttack\cite{fang2024unified} indiscriminately accumulate millions of low-quality samples. This not only introduces privacy risks and annotation noise but also leads to rapid performance saturation. However, constructing large-scale image-aligned textual data with Chain-of-Thought (CoT) reasoning remains a significant challenge for CPT methods. Recent surveys\cite{zheng2026unified} have consistently highlighted that current research generally lacks a unified system capable of achieving deep cross-attack alignment at the semantic-level while simultaneously ensuring security auditability and engineering feasibility. This fundamental deficiency severely restricts the trustworthy deployment of MLLMs in real-world security applications. 

To address this issue, researchers have primarily proposed two approaches: augmenting the model with external tools to enrich its semantic understanding of forgery artifacts, or incorporating CoT reasoning during post-training. In terms of tool-augmented reasoning, Zhang et al. have proposed Tool-Augmented Reasoning for Face Anti-Spoofing (TAR-FAS)\cite{zhang2026intuition} allow models to invoke external operators for forensic analysis, this plug-in architecture expands the attack surface of FAS systems. However, once the external tools fail or are subjected to adversarial perturbations, the end-to-end security is immediately compromised. Given the inherent limitations of external tool invocation, such as increased inference latency and dependency on tool availability, it becomes particularly crucial to enhance the intrinsic reasoning capabilities of the model itself, enabling it to autonomously identify and reason about forgery artifacts without external assistance.

\begin{figure}[htbp]
    \centering
    \includegraphics[width=0.95\linewidth]{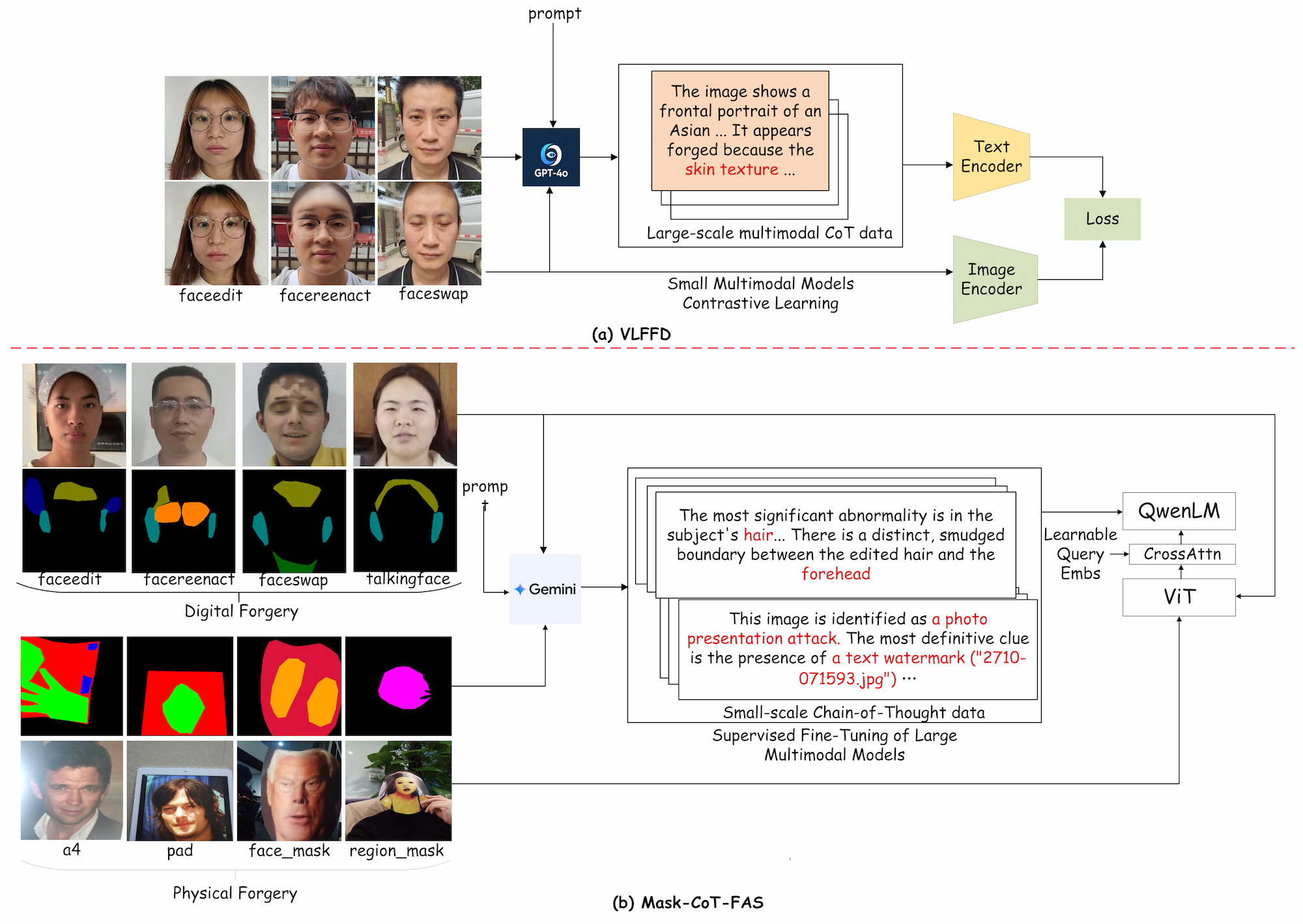}
    \caption{VLFFD proposes a CoT construction method; however, it is applicable only to Deepfake attacks with precise forgery masks, and the method itself inherently follows the CPT paradigm, making it difficult to eliminate the reliance on large-scale image-text paired data. In contrast, our proposed data and framework are applicable to all attack types. As a post-training approach, it elicits the reasoning capabilities of MLLMs with only a small amount of data.}
    \label{fig:intro}
\end{figure}

In the realm of enhancing the explainability of large models for anti-spoofing, researchers have already made explorations. Ke Sun et al. proposed a fine-grained\cite{sun2025towards}, instance-level prompt generation method, demonstrating the importance of spatial information in mitigating hallucinations during CoT generation. However, their approach requires pixel-level matched Deepfake attacks, making it difficult to obtain such precise annotations for every image in physical attack scenarios. Wang et al. relied on the prompt-guided visual token masking strategy\cite{wang2026faceshield}. As a model training technique, prompt-guided vision token masking forces the model to focus on easily confused attack regions through random visual token masking to improve generalization. Nevertheless, this approach remains at a relatively macroscopic or coarse-grained stage, a major limitation of this approach is that it is exclusively applicable to physical attacks.

To address the aforementioned security capability gaps, this paper proposes MFAD, an explainable reasoning system for UFAD, accompanied by a semantic-level annotation benchmark.  Driven by the practical demands of security scenarios, we redesign a fine-grained pixel-semantic anchoring mechanism.  It remains challenging to unify physical and digital attacks at the semantic level, thereby hindering the construction of a truly generalized face anti-spoofing large model. We utilize semantic-level annotations to unify physical and digital attacks, enabling robust CoT reasoning without strict pixel alignment  as shon in Figure \ref{fig:intro} (b). We abandon the reliance on external tools and the blind pursuit of massive low-quality data, opting instead to directly activate the intrinsic anti-spoofing reasoning capabilities of LLMs through refined semantic annotations, thereby ensuring the auditability and anti-interference of the reasoning paths. Specifically, we construct a unified evaluation benchmark encompassing 16 categories of attacks and pioneer a cross-attack semantic-level unified annotation paradigm: we capture local forgery traces by segmenting 12 semantic regions for deepfakes; annotate moiré patterns and illumination anomalies for display replay attacks; and precisely annotate carrier attributes for mask attacks. Following a cost-quality trade-off, we annotate only 1,000 semantic masks per attack category. Through a quality control process involving dual-person verification and ambiguity arbitration, we generate reasoning evidence chains that strictly correspond to the spoofed regions, fundamentally circumventing localization hallucinations.

We conducted comprehensive system validation on mainstream MLLM foundations. Experimental results demonstrate that MFAD, approximately half of the total 16,000 samples, achieves about 40\% to 50\% relative reduction in in-domain Average Classification Error Rate (ACER) and restricts cross-domain performance degradation to within 11.62\%/5.23\%, significantly outperforming existing methods. Under Projected Gradient Descent with 10 iterations (PGD-10) and AutoAttack white-box attacks, the detection accuracy decreases by only 3.2\%, validating the intrinsic defense capability of the semantic anchoring mechanism against adversarial perturbations. Supplementary evaluations by domain practitioners show that the reasoning evidence chains generated by the model receive an average score of 4.5734/5 for readability and evidence reliability, and the system's inference latency satisfies the deployment requirements of real-time security applications. These results fully demonstrate that the few-shot, high-quality semantic annotation paradigm is an effective approach for building trustworthy, explainable, and cost-effective UFAD systems, providing a novel engineering paradigm for the deployment of MLLMs in the security domain.

Our main contributions can be summarized as follows:\\
(1) We comprehensively survey 16 types of physical and digital face attacks and construct 1,000 fine-grained, semantic-level CoT data samples for each category. This provides the community with a large-scale, high-quality dataset featuring precise textual descriptions, thereby laying a solid foundation for research in UFAD.\\
(2) By training 3B and 7B parameter models on a subset of the proposed dataset, we develop a compact face anti-spoofing large model capable of simultaneously defending against both physical and digital attacks. Remarkably, this model outperforms most existing MLLMs with significantly larger parameter scales, while demonstrating robust cross-domain generalization capabilities. \\
(3) During our experiments, we observed a significant trade-off between robustness and generalization when MLLs are applied to face anti-spoofing tasks, and we explore this issue through Reinforcement Learning (RL).
\section{Related Work}
In addition to the state-of-the-art methods discussed above, we first supplement our review with earlier multimodal approaches. Traditional CNN-based methods are constrained by limited feature expressiveness, typically addressing only certain types of spoofing attacks. Relying on multiple specialized models in practical deployment increases computational burden and undermines system reliability, making a unified model for all spoofing types a desirable future direction. We then provide an in-depth discussion of UFAD as a precursor to MFAD. Finally, since we also construct a dataset in this work, we briefly survey existing datasets in the face anti-spoofing domain.
\subsection{VLM and MLLM-based Face Anti-Spoofing}
The introduction of Vision-Language Model (VLM) has brought new opportunities to FAS. Fast Language-Image Pre-training (FLIP)\cite{srivatsan2023flip} bypasses the need for complex additional modules by leveraging the zero-shot capabilities of Contrastive Language-Image Pre-training (CLIP)\cite{radford2021learning}. It aligns image features with a set of natural language-based class descriptions. This paper demonstrates that in low-data scenarios, utilizing linguistic semantics alone as an anchor can significantly reduce the feature distance between the source and target domains, thereby achieving robust cross-domain generalization. To address the misalignment issue of coarse semantics versus fine-grained textures, Fang et al. introduce a dual-branch alignment strategy\cite{fang2024vl}. One branch aligns global semantics, while the other employs textual prompts to guide the model's attention toward local forgery cues, distilling this fine-grained attention back into a lightweight visual encoder. Peng et al. proposes a bidirectional visual-language fusion network\cite{peng2025mllm}. Unlike simple prompting approaches, it incorporates a bidirectional interaction module at the feature level, enabling textual features to directly participate in refining visual features, and vice versa. Furthermore, they constructed a companion dataset, which provides fine-grained textual descriptions of forgery types and artifacts for each deepfake frame.

\subsection{Unified Physical-Digital Face Anti-Spoofing}
As attack vectors evolve from isolated physical presentation attacks (e.g., printed photographs, 3D masks) to complex digital tampering (e.g., deepfake, faceswap) and hybrid attacks, constructing a unified detection framework capable of simultaneously defending against both threats has become a prominent research focus in academia. To address the significant discrepancies in feature distributions between physical and digital attacks, several studies have focused on bridging the feature gap through data-level interventions. He et al. \cite{he2024joint} proposed a joint physical-digital attack detection framework based on the core idea of algorithmically simulating the common cues shared by both attack types. Rather than treating physical and digital attacks as independent domains, this method forces the network to learn universal forgery representations by generating synthetic data that incorporates characteristics of both attack types.

To bridge the gap between physical and digital attacks, the texture-semantic dual-track Unification framework proposes a unified architecture\cite{zhang2025generalizable}. One track utilizes traditional CNNs/ViTs to extract high-frequency texture features, while the other leverages CLIP to extract semantic features; these are subsequently fused to yield a unified score. The study demonstrates that CLIP semantic features exhibit inherent robustness against semantic attacks, such as 3D masks, whereas texture features are more sensitive to digital synthesis attacks, proving that their complementarity constitutes the ultimate solution.

To accommodate the substantial disparities between physical and digital attacks, researchers have adopted more flexible network architectures to enable models to dynamically select processing pathways. Liu et al. utilizes a dual cross-attention mechanism and a semi-fixed Mixture-of-Experts (MoE) model to achieve generalization across diverse attack types\cite{liu2024moeit}. Similarly, Zou et al. \cite{zou2024softmoe} introduced La-SoftMoE CLIP, which effectively improves the performance of CLIP models in unified detection tasks by allocating expert resources via a soft gating mechanism. Chen et al. \cite{chen2024facecat} proposed FaceCat, a framework based on a unified generative model. It aims to simultaneously capture the deep-seated features of both physical and digital attacks through variants of Generative Adversarial Networks (GANs), thereby enhancing the security of facial recognition systems. Leveraging the generalization capabilities of large-scale pre-trained models represents another major trend in addressing cross-domain detection. 

Through image-text alignment pre-training, the model leverages semantic information to assist in decision-making, demonstrating superior performance on unseen attack types. Beyond model architectures, optimization strategies during the training process are equally critical. Yu et al. \cite{yu2024unified} designed a two-stage training strategy that involves pre-training on massive general-purpose data followed by fine-tuning on specific domain data, effectively mitigating overfitting. Focusing on the visual analysis of data domain shifts within CNNs, Huang et al. \cite{huang2024visualization} proposed an optimized method for selecting classification thresholds. This is of significant practical importance for balancing the false acceptance rates of physical and digital attacks during real-world deployment.

Although the aforementioned works (\cite{deb2023unified}\cite{he2024joint}\cite{yu2024unified}\cite{liu2024moeit}) have achieved remarkable progress in unified detection, existing methods still face two primary challenges: lack of explainability and  insufficient fine-grained alignment. Most existing models (\cite{chen2024facecat}\cite{zou2024softmoe}) remain black-box classifiers, failing to provide specific forgery evidence (e.g., this is a screen moiré pattern or this is the material boundary of a mask). Insufficient Fine-Grained Alignment: Existing VLM-based methods often rely on global image features and lack pixel-level semantic anchoring, making them prone to localization hallucinations.

\subsection{Construction and Evolution of Anti-Spoofing Datasets}
Data serves as the foundation for training anti-spoofing models, yet current dataset construction primarily faces the contradiction between massive low-quality data and scarce fine-grained data. Commonly used academic datasets, such as FaceForensics++ (FF++)\cite{rossler2019faceforensics++} and Celeb-DF\cite{li2020celeb}\footnote{CelebA-DF (Celeb-DeepFake), a large-scale dataset for DeepFake detection featuring high-quality synthesized faces with fewer visual artifacts.}, mainly target specific digital generators and suffer from data bias; FF++ focuses on four common manipulation methods to benchmark deepfake detection, while Celeb-DF improves upon visual quality to reduce artifacts and challenge detection algorithms. In contrast, data for physical attacks often incurs high costs for custom collection, necessitating datasets like CASIA-SURF\cite{zhang2020casia}\footnote{CASIA-SURF is a large-scale multi-modal face anti-spoofing dataset introduced by Institute of Automation, Chinese Academy of Sciences (CASIA). The term SURF refers to Speeded Up Robust Features, a widely used feature detection and description method in computer vision.} and Wide Multi-Channel presentation Attack (WMCA)\cite{george2019biometric}, which introduce multi-modal channels (Color, Depth, Infrared) to capture material properties, with WMCA further expanding this to include thermal data for robust 3D mask detection. Addressing the scarcity of fine-grained annotations, CelebA-Spoof provides a large-scale collection with rich diversity across 8 scenes and over 10 camera sensors\cite{zhang2020celeba}, offering detailed spoofing type and attribute annotations to support comprehensive analysis. Meanwhile, Multi-dimensional Face Forgery Image (MFFI) tackles the challenge of complex, multi-modal forgeries by providing data specifically designed to test the robustness of models against sophisticated\cite{miao2025mffi}, combined manipulation attacks. To further bridge the gap between controlled lab settings and real-world deployment. MS-UFAD incorporates detailed textual descriptions for attack clues\cite{jiang2025ms}—the first of its kind—alongside 795k videos and 60k images covering 52 attack methods, enabling text-guided detection methods to leverage semantic assistance for significantly improved accuracy. Although MS-UFAD is a large-scale image-text dataset capable of joint physical and digital face anti-spoofing, it still suffers from several significant limitations: (1) The textual descriptions are overly brief and contain numerous errors; (2) It lacks explicit forgery cues; and (3) It covers a limited variety of physical spoofing attacks, particularly lacking mask-based attacks.

\section{Method}
We first theoretically examine UFAD based on MLLMs, along with its existing challenges. (Section 3.1). Subsequently, we introduce the core fine-grained pixel-semantic anchoring mechanism, which explicitly binds visual forgery cues to the reasoning chain (Section 3.2). Next, based on the data construction workflow, we comprehensively describe the pipeline for building the cross-attack semantic-level unified annotation benchmark, along with the multi-level quality control strategies (Section 3.3). Finally, we present the supervised fine-tuning scheme based on multimodal large language models (Section 3.4).
\subsection{Problem Formulation}
In the field of UFAD based on MLLM, the raw input consist of an face image $I$ and text $T$.  $T$ denotes the textual query or prompt, which is used to instruct the large model to analyze image characteristics and subsequently provide a conclusion regarding its authenticity. Feature extraction is performed via modality encoders trained on face anti-spoofing datasets (such as a visual encoder $\mathcal{E}_v$ and a text encoder $\mathcal{E}_t$). The image and text feature vectors can be denoted as $F_v = \mathcal{E}_v(I; \theta_v)$ and $F_t = \mathcal{E}_t(T; \theta_t)$, respectively.  $\theta_v$ and $\theta_t$ denote the encoder parameters. These parameters are typically frozen during the reasoning phase. By employing a linear projector $\mathcal{P}$ to align visual features with the text embedding space. The model leverages the attention mechanism to achieve dynamic cross-modal information interaction. $F_v$, $F_t$, and $P_v$ are mapped to $Q$, $K$, and $V$, respectively. The fused multimodal content can be denoted as $H_{\text{fused}}$. 

The fused multimodal context $H_{\text{fused}}$ is concatenated with the original text instruction and injected as a conditional input into the Large Language Model (LLM) for reasoning. The inference process of the LLM is strictly modeled as the joint probability factorization of the target output sequence given the multimodal context. During the inference stage, the model employs strategies such as greedy decoding or beam search to find the next token that maximizes the conditional probability at each time step $t$:

\begin{equation}
\hat{y}_t = \arg \max_{y_t} P(y_t \mid y_{<t}, H_{\text{fused}}, T; \Theta)
\end{equation}

where $y_{<t}$ refers to all tokens generated before time step $t$, denoted as $\{y_1, \dots, y_{t-1}\}$. This ensures the coherence and grammatical correctness of the generated content. $T$ refers to the original text instruction. Serving as a system prompt or user query, it remains constant throughout the generation process, providing the task context for reasoning. $\Theta$ refers to the model parameters.

From the perspective of computational complexity, a single forward pass of the transformer architecture has a fixed computational depth\cite{sanford2023representational}, making it inherently insufficient for problems requiring multi-step logical reasoning. CoT essentially trades sequence length for computational depth: each generated intermediate token $z_t$ serves as a complete nonlinear state update, enabling the model to construct a smooth geodesic through a high-dimensional feature space where decision boundaries are otherwise rugged and non-convex. In high-fine-grained tasks such as face anti-spoofing, where genuine and forged images heavily overlap in feature space, short-text fine-tuning lacks such intermediate states $Z$, forcing the model to search for the global optimum directly on a rugged loss landscape\cite{jiang2025ms, fang2024unified}, which readily leads to gradient degradation or decision boundary collapse.

The prevailing paradigm for training UFAD models relies on post-training with powerful foundation models such as CLIP or Qwen-VL\cite{Qwen-VL}, yet the primary bottleneck remains acquiring high-quality image-text aligned data. Existing remedies including knowledge distillation, offline CoT generation, and large-scale short-text fine-tuning\cite{jiang2025ms}---inevitably introduce varying degrees of hallucination\cite{zheng2026unified}. To address this, we constrain the language decoding phase with a structured three-stage CoT template (Observation $\rightarrow$ Attribution $\rightarrow$ Judgment), in which each evidence chain must be explicitly grounded in pixel-level masks. These masks serve as hard spatial anchors that impose a localization-analysis-judgment paradigm on CoT generation, fundamentally curbing hallucinations by ensuring every reasoning step is tied to verifiable visual evidence rather than open-ended, self-reinforcing textual patterns.

\subsection{Chain-of-Thought Evidence Generation}
The construction of high-quality semantic anchoring data serves as the cornerstone of the MFAD system. Most existing generation paradigms adopt a Large Language Model (LLM)-assisted approach to synergize the strengths of human experts and current proprietary LLMs. Relying entirely on humans for CoT annotation is prohibitively labor-intensive and error-prone, whereas delegating the task solely to LLMs often results in a failure to capture critical focal points and induces hallucinations\cite{huang2024ffaa}. To mitigate these issues, this paper employs human experts to annotate masks as guidance for the models, thereby fully leveraging the capabilities of LLMs in analyzing subtle, pixel-level features that are imperceptible to the naked eye. Section \ref{sec:slmg} details the mask generation process, and finally, Section \ref{sec:uslab} presents the overall data generation pipeline.
\subsubsection{Semantic-level Joint Mask Generation}\label{sec:slmg}
To facilitate precise CoT reasoning and fine-grained feature learning, we formulate the data construction process as a semantic mapping problem. Let $\mathcal{D} = \{(I_i, M_i)\}_{i=1}^N$ denote the constructed dataset comprising $N$ samples, where $I_i \in \mathbb{R}^{H \times W \times 3}$ represents the input image and $M_i \in \mathbb{R}^{H \times W \times 3}$ denotes the corresponding pixel-level annotation mask.

\begin{figure}[htbp]
    \centering
    \includegraphics[width=0.95\linewidth]{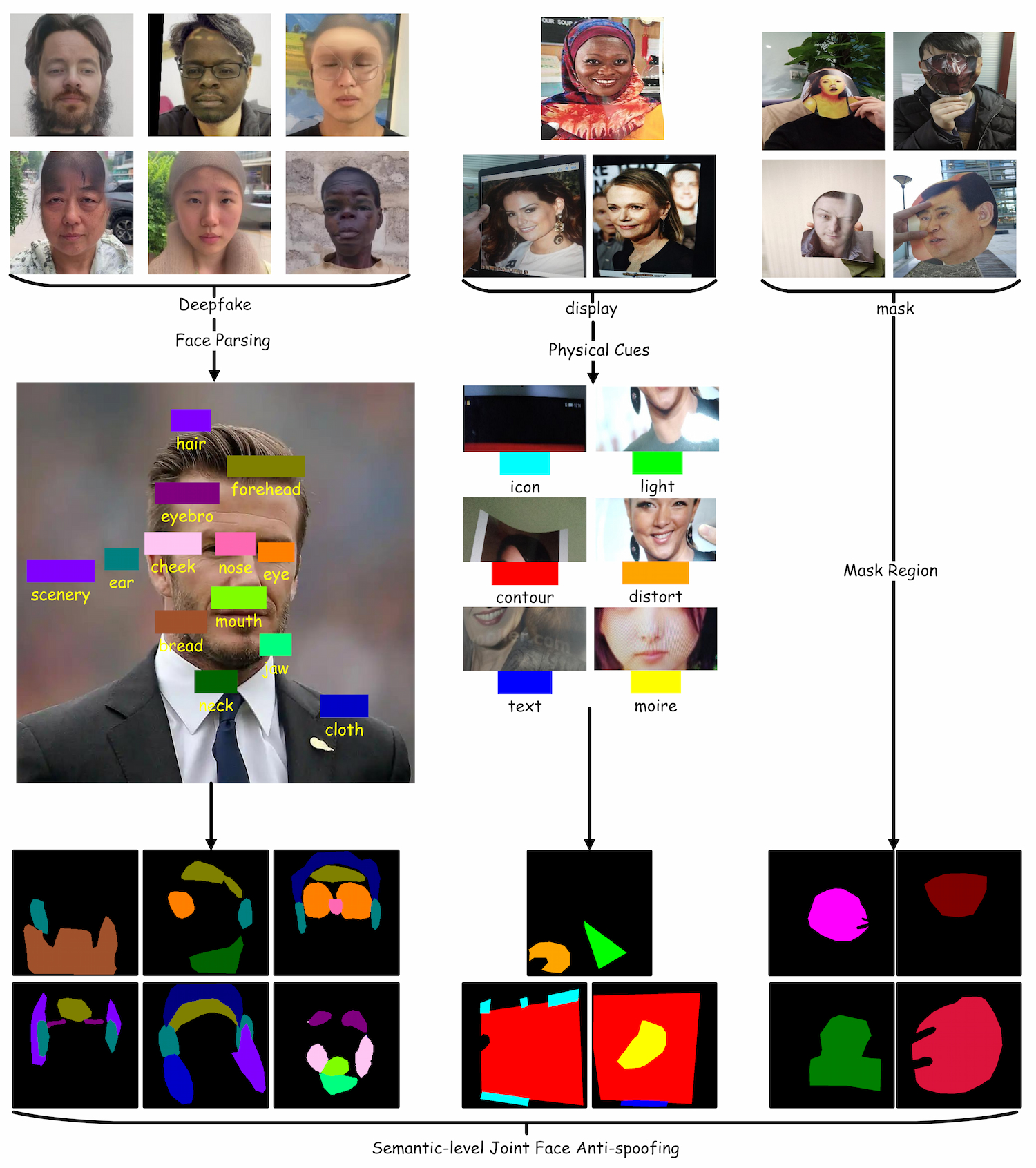}
    \caption{Schematic diagram of the mask generation process.}
    \label{fig:dataset_generation_pipline-v3}
\end{figure}

Unlike conventional binary supervision, we define a color-coded semantic annotation protocol. We establish a finite set of forgery trace categories $\mathcal{C}$ and a bijective mapping function $\Phi : \mathcal{C} \to \mathbb{Z}^3$ that assigns a unique RGB pixel value to each semantic category. The set of all annotated semantic regions $\mathcal{C}$ is partitioned into three disjoint subsets based on the attack modality: $\mathcal{C}_{\text{deepfake}}$, $\mathcal{C}_{\text{display}}$, and $\mathcal{C}_{\text{mask}}$. For digital forgeries generated via face editing, reenactment, or swapping, artifacts are localized in specific facial components. We define this subset as:
\begin{equation}
\begin{split}
\mathcal{C}_{\text{deepfake}} = \{&c_{\text{hair}}, c_{\text{forehead}}, c_{\text{eyebrow}}, c_{\text{eye}}, c_{\text{ear}}, c_{\text{nose}},\\ 
&c_{\text{cheek}}, c_{\text{mouth}}, c_{\text{jaw}}, c_{\text{beard}}, c_{\text{neck}}, c_{\text{cloth}}, c_{\text{scenery}}\}
\end{split}
\end{equation}
where each element corresponds to a parsed facial region exhibiting synthesis traces.

\textbf{Physical Presentation Cues ($\mathcal{C}_{\text{display}}$):} 
For photo display attacks (e.g., printed photos, screens), we annotate physical artifacts rather than semantic parts. This subset is defined as:
\begin{equation}
\mathcal{C}_{\text{display}} = \{c_{\text{contour}}, c_{\text{light}}, c_{\text{text}}, c_{\text{moire}}, c_{\text{distort}}, c_{\text{icon}}\}
\end{equation}
representing boundary reflections, screen moiré patterns, text overlays, and geometric distortions.

\textbf{Mask Attack Regions ($\mathcal{C}_{\text{mask}}$):} 
For physical mask attacks, the annotation targets the occluding object itself:
\begin{equation}
\mathcal{C}_{\text{mask}} = \{c_{\text{face\_mask}}, c_{\text{region\_mask}}, c_{\text{three\_d\_mask}}, c_{\text{upper\_body\_mask}}\}
\end{equation}

To enable the model to distinguish these categories implicitly through color channels during CoT generation, we define the mapping $\Phi(c)$ for any category $c \in \mathcal{C}$. The ground truth mask $M_i$ at pixel location $(x, y)$ is generated by applying the mapping function $\Phi$ to the semantic label map. Formally, this is expressed as:
\begin{equation}
M_i(x, y) = 
\begin{cases} 
\Phi(c) & \text{if } (x, y) \in \Omega_c \\
(0, 0, 0) & \text{otherwise (Background)}
\end{cases}
\end{equation}
where $\Omega_c$ denotes the spatial region of category $c$ in image $I_i$. The specific encoding values are defined as follows (subset representation):

\noindent For Display Cues:
\noindent For Display Cues:
\begin{equation}
\begin{aligned}
    \Phi(c_{\text{contour}}) &= (255, 0, 0)_{\text{Red}}, & \Phi(c_{\text{light}}) &= (0, 255, 0)_{\text{Green}}, \\
    \Phi(c_{\text{text}}) &= (0, 0, 255)_{\text{Blue}}, & \Phi(c_{\text{moire}}) &= (255, 255, 0)_{\text{Yellow}}, \\
    \Phi(c_{\text{icon}}) &= (0, 255, 255)_{\text{Cyan}}, & \Phi(c_{\text{distort}}) &= (255, 165, 0)_{\text{Orange Red}}.
\end{aligned}
\end{equation}

\noindent For Mask Regions:
\begin{equation}
\begin{aligned}
    \Phi(c_{\text{region\_mask}}) &= (255, 0, 255)_{\text{Magenta}}, \\
    \Phi(c_{\text{three\_d\_mask}}) &= (128, 0, 0)_{\text{Dark Red}}, \\
    \Phi(c_{\text{upper\_body\_mask}}) &= (0, 128, 0)_{\text{Dark Green}}, \\
    \Phi(c_{\text{face\_mask}}) &= (220, 20, 60)_{\text{Crimson}}.
\end{aligned}
\end{equation}

\noindent For Deepfake Facial Parts (Selected Examples):
\begin{equation}
\begin{aligned}
    \Phi(c_{\text{hair}}) &= (0, 0, 128)_{\text{Dark Blue}}, & \Phi(c_{\text{eye}}) &= (255, 128, 0)_{\text{Orange}}, \\
    \Phi(c_{\text{mouth}}) &= (128, 255, 0)_{\text{Chartreuse}}, & \Phi(c_{\text{nose}}) &= (255, 105, 180)_{\text{Hot Pink}}.
\end{aligned}
\end{equation}

This formalization ensures that the annotation space $\mathcal{M}$ is discrete and semantically structured, providing a rigorous supervisory signal for multi-class forgery localization.

\subsubsection{Unified Semantic-Level Annotation Benchmark}\label{sec:uslab}
After obtaining the masks, we construct fine-grained CoT data based on both the masked images and the original images. Let $\mathcal{D} = \{(I_i, M_i)\}_{i=1}^N$ denote the dataset construction process, where $I$ represents the input image and $M$ represents the corresponding pixel-level ground truth mask. To address the scalability bottleneck of manual annotation, we define an automated generation function $\mathcal{F}_{gen}$ driven by a LLM (e.g., Gemini). The generation process for a single sample is defined as:
\begin{equation}
    R_{raw} = \mathcal{F}_{gen}(I, M, \mathcal{P}_{sys}, \mathcal{K}_{aux})
\end{equation}
where $R_{raw}$ is the raw generated response, $\mathcal{P}_{sys}$ represents the system prompts, and $\mathcal{K}_{aux}$ denotes the auxiliary knowledge base comprising category and trace definitions. To ensure data quality, we employ a Tri-Stage Quality Assurance Protocol. First, in the Model-Based Evaluation stage, two independent evaluator models assign quality scores $S_A, S_B \in [0, 1]$ to the Chain-of-Thought (CoT) reasoning in $R_{raw}$. Second, the Decision Logic dictates that samples are accepted if $S_A > \tau$ and $S_B > \tau$ (where $\tau$ is a high-score threshold), rejected if both scores are below the threshold, and routed to human experts for revision if the scores diverge (i.e., $(S_A > \tau \land S_B \leq \tau) \lor (S_A \leq \tau \land S_B > \tau)$). Finally, the validated response is structured into a Visual Question Answering (VQA) tuple: $D_{final} = (Q, A, I)$.

\begin{figure}[htbp]
    \centering
    \includegraphics[width=5cm]{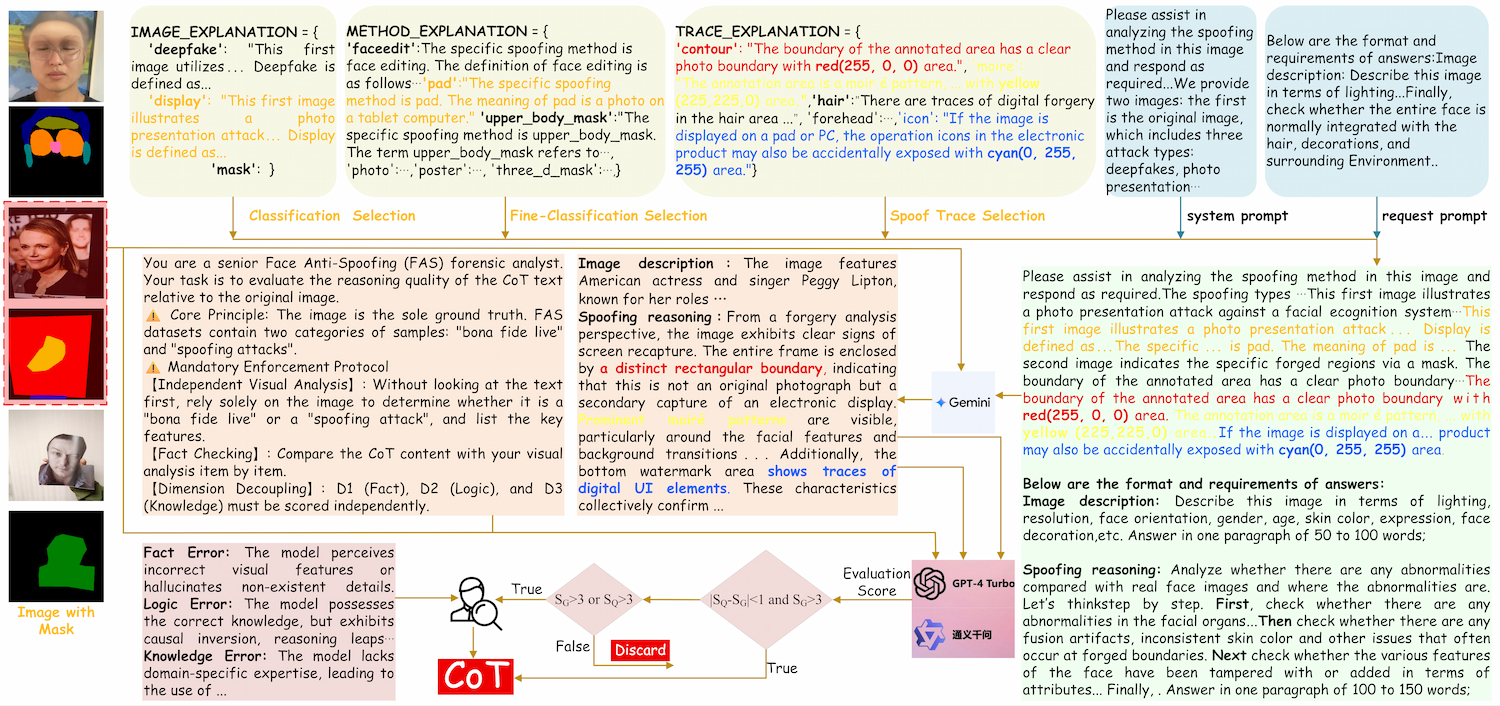}
    \caption{CoT Data Construction Pipeline.}
    \label{fig:dataset_generation_pipline-v5}
\end{figure}

To guide the model in open-world face forgery analysis, we construct a composite prompt space $\mathcal{P}_{total}$. The prompt is structured into five distinct components:
\begin{equation}
    \mathcal{P}_{total} = \{ \mathcal{P}_{cat}, \mathcal{P}_{sub}, \mathcal{P}_{trace}, \mathcal{P}_{role}, \mathcal{P}_{cot} \}
\end{equation}
Here, Category Selection ($\mathcal{P}_{cat}$) defines the coarse-grained label space $\mathcal{C}_{coarse} = \{ \text{Deepfake}, \text{Display}, \text{Mask} \}$. Fine-grained Selection ($\mathcal{P}_{sub}$) defines the specific forgery types, where $|\mathcal{C}_{sub}^{df}|=4$, $|\mathcal{C}_{sub}^{disp}|=6$, and $|\mathcal{C}_{sub}^{mask}|=4$. Forgery Trace Selection ($\mathcal{P}_{trace}$) comprises a comprehensive set of 23 forensic traces $\mathcal{T} = \{t_1, \dots, t_{23}\}$, characterized by both spatial and physical properties. The System Prompt ($\mathcal{P}_{role}$) defines the agent's role, boundary constraints, and behavioral norms, while the CoT Requirement ($\mathcal{P}_{cot}$) enforces step-by-step reasoning to enhance logical consistency. During generation, the input to the model includes the visual features and explicit pixel coordinates derived from $M$ to ground the analysis spatially.

Subsequently, the generated prompt, the original face image, and the mask are provided as inputs to the Gemini large model. The reasoning capability of large MLLMs is highly dependent on the attention mechanism of transformers. During the testing phase, the mask $M$ directly alters the model's attention weight distribution in the analysis phase. Let the visual attention score of the model at step $t$ be $e_t$. After Softmax normalization, the attention distribution $A_t$ is obtained:
\begin{equation}
    A_t = \text{softmax} \left( \frac{Q_t K_V^T}{\sqrt{d}} \right)
\end{equation}

When the input includes a mask $M$, Gemini 3.1's attention mechanism implicitly or explicitly uses $M$ as a filtering condition, causing a drastic change in the attention distribution $A'_t$:
\begin{equation}
    A'_t = \text{softmax} \left( \frac{Q_t (K_V \odot M)^T}{\sqrt{d}} \right)
\end{equation}

During the testing phase, the $\odot$ operation (or an equivalent masked attention mechanism) is equivalent to performing a \textbf{Hard Masking} on the global visual key. When reasoning, the model's gaze is forcibly locked within the masked region. Since hallucinations often occur because the model looks in the wrong place' or 'over-generalizes the background', this redistribution of attention during testing physically cuts off the visual input source for hallucination generation at the physical level.

To curate high-quality reasoning samples from the massive generated pool, we design a dual-model consensus-based adaptive filtering mechanism. Let $S_{GPT}$ and $S_{Qwen}$ denote the evaluation scores assigned by GPT-4o and Qwen to a specific CoT sample $R_{raw}$, respectively. We define a consistency threshold $\delta = 1$ and a quality threshold $\tau = 3$. The filtering logic is formalized as a piecewise decision function \begin{equation}
    \begin{split}
        \Phi(R_{raw}) = 
        \begin{cases} 
        \text{Accept}, & \text{if } S_{GPT} > \tau \land S_{Qwen} > \tau \land |S_{GPT} - S_{Qwen}| < \delta \\
        \text{Review}, & \text{if } |S_{GPT} - S_{Qwen}| \ge \delta \land \max(S_{GPT}, S_{Qwen}) > \tau \\
        \text{Reject}, & \text{otherwise}
        \end{cases}
    \end{split}
\end{equation}
This mechanism effectively balances automation efficiency with the recall rate of long-tail samples. Samples falling into the ``Human-Review'' category are routed to the refinement stage, ensuring that potentially valuable but controversial reasoning paths are not prematurely discarded.

For samples routed to the human stage, we propose a fine-grained refinement framework based on error diagnosis. This process is modeled as a mapping transformation $\Psi: R_{raw} \to R_{gold}$, converting a defective reasoning trajectory into a gold-standard one. Human experts first perform an Error Attribution Diagnosis $\mathcal{D}_{iag}$, decoupling the error space into three orthogonal categories: Factual Errors ($E_{fact}$, e.g., visual hallucinations), Logical Errors ($E_{logic}$, e.g., reasoning jumps), and Knowledge Errors ($E_{know}$, e.g., lack of domain expertise). Subsequently, targeted rewriting operations $\mathcal{O}_{p}$ are executed based on the diagnosis:
\begin{itemize}
    \item \textbf{Feature Alignment:} For $E_{fact}$, experts correct the visual descriptions to align with the ground truth image.
    \item \textbf{Logic Interpolation:} For $E_{logic}$, intermediate reasoning steps are inserted to bridge logical gaps (e.g., inferring $B$ from $A$ to reach $C$).
    \item \textbf{Knowledge Injection:} For $E_{know}$, domain-specific definitions and criteria are injected to replace incorrect assumptions.
\end{itemize}
The final output $R_{gold}$ ensures strict consistency among visual evidence, logical chains, and domain knowledge. The overall procedure for CoT filtering and refinement is summarized in Algorithm \ref{alg:cot_filtering}.

\begin{algorithm}[t]
\caption{CoT Filtering and Human Refinement Protocol}
\label{alg:cot_filtering}
\KwIn{Raw CoT sample $R_{raw}$, Image $I$, Thresholds $\tau=3, \delta=1$}
\KwOut{Final CoT sample $R_{final}$}

$S_{GPT} \leftarrow \text{Evaluate}(R_{raw}, \text{GPT-4o})$\;
$S_{Qwen} \leftarrow \text{Evaluate}(R_{raw}, \text{Qwen})$\;
$\Delta_{score} \leftarrow |S_{GPT} - S_{Qwen}|$\;

\eIf{$S_{GPT} > \tau$ \textbf{and} $S_{Qwen} > \tau$ \textbf{and} $\Delta_{score} < \delta$}{
    $R_{final} \leftarrow R_{raw}$\; \tcp*{Auto-Accept}
}{
    \eIf{$\Delta_{score} \ge \delta$ \textbf{and} $\max(S_{GPT}, S_{Qwen}) > \tau$}{
        $E_{type} \leftarrow \text{DiagnoseError}(R_{raw}, I)$\; \tcp*{Human Diagnosis}
        $R_{final} \leftarrow \text{HumanRewrite}(R_{raw}, E_{type})$\; \tcp*{Rewriting \& Interpolation}
    }{
        $R_{final} \leftarrow \text{NULL}$\; \tcp*{Reject}
    }
}
\KwRet{$R_{final}$}
\end{algorithm}

Ultimately, the 16,000 samples that passed the three-level verification constitute the semantic anchoring training set for Mask-CoT-FAS. In terms of sample size, this dataset is merely 1.5\%--3\% of existing large-scale FAS instruction-tuning datasets (e.g., FAS-MLLM, LLaFA). However, it achieves an order-of-magnitude improvement in pixel-semantic alignment density and reasoning evidence reliability, providing a solid foundation for subsequent few-shot efficient fine-tuning.

\subsection{Training Process and Optimization Strategy}
We formulate the post-training of our unified face anti-spoofing model as a constrained optimization problem, aiming to align the model with human reasoning intent while jointly detecting both physical attacks (e.g., printed photos, 3D masks) and digital forgeries (e.g., Deepfake face-swapping).
\subsubsection{Training Objective}
We perform Supervised Fine-Tuning (SFT) with a standard auto-regressive cross-entropy loss, while constraining the model within a KL trust region to prevent reward hacking\cite{ouyang2022instructgpt}. The training objective is defined as:

\begin{equation}
\min_\theta \mathcal{D}(P_{target}(Y|X) \parallel P_\theta(Y|X)) + \lambda \mathcal{R}(\theta)
\end{equation}

where $\mathcal{R}(\theta)$ is the regularization term and $\lambda$ is the regularization coefficient. Since $P_{target}$ is typically unknown and sparse, direct optimization risks reward hacking---the model deviating from its pre-trained representations. We therefore formulate post-training as a constrained optimization problem, restricting the model within a KL trust region centered on the reference policy $\pi_{ref}$:

\begin{equation}
    \max_{\pi_\theta} \mathbb{E}[\mathcal{J}(X, Y)] \quad \text{s.t.} \quad \mathbb{E}_X [\text{KL}(\pi_\theta(\cdot|X) \parallel \pi_{ref}(\cdot|X))] \leq \epsilon
\end{equation}

Concretely, we perform Supervised Fine-Tuning (SFT) using the standard auto-regressive cross-entropy loss over the token sequence $y = \{y_1, \ldots, y_T\}$:

\begin{equation}
    \mathcal{L}(\theta) = -\sum_{t=1}^{T} \log P(y_t | y_{<t}, x; \theta)
\end{equation}

where $x$ is the input query and $\theta$ denotes the trainable LoRA parameters. This objective trains the model to generate structured CoT reasoning that jointly identifies physical attacks (e.g., printed photos, 3D masks) and digital forgeries (e.g., Deepfake face-swapping), enabling unified face anti-spoofing.

\subsubsection{LoRA Configuration and Capacity Justification}
To achieve parameter-efficient adaptation while preserving the model's pre-trained generalization capability, we employ Low-Rank Adaptation (LoRA) with a theoretically grounded rank selection\cite{hu2021lora}. Rather than updating the full parameter set, we freeze the pre-trained weights and inject trainable low-rank decomposition matrices, whose sufficiency is justified by the low intrinsic dimension of reasoning tasks and the data-parameter efficiency constraint. The forward pass for a given layer is formulated as:

\begin{equation}
    h = W_0 x + BAx
\end{equation}

where $A \in \mathbb{R}^{r \times k}$ and $B \in \mathbb{R}^{d \times r}$ with rank $r=8$ and scaling factor $\alpha=32$. LoRA adapters are applied to all linear layers.

We justify the sufficiency of this low-rank configuration from two perspectives. First, recent studies show that fine-tuning pre-trained models on downstream tasks operates within a low-dimensional subspace, where the intrinsic dimension $d_{int} \ll |\theta_{full}|$. Since CoT reasoning introduces logical alignment rather than massive new factual knowledge, the required parameter update is sparse in the high-dimensional space, and the low-rank decomposition $\Delta W \approx \sum_{i=1}^{r} \sigma_i u_i v_i^T$ is sufficient to capture the necessary variance. Second, from a statistical learning perspective, full fine-tuning of a 7B model ($\sim 7 \times 10^9$ parameters) on $N=20{,}000$ CoT samples yields an extreme over-parameterization ratio of $\approx 3.5 \times 10^5$ parameters per sample, posing severe overfitting and catastrophic forgetting risks. LoRA reduces the trainable parameters to approximately $0.1\%$--$0.5\%$ of the total, striking a balance between expressiveness and regularization---sufficient to model the shift in $P(y|x, \text{CoT})$ while preserving the robust representations learned during pre-training.

\section{Experimental Evaluation}
The experimental evaluation is divided into three parts: experimental setup, ablation studies, and comparative experiments, which are detailed in Sections \ref{sec:es}, \ref{sec:as} and \ref{sec:ce}, respectively.
\subsection{Experimental Setup}\label{sec:es}
\subsubsection{Evaluation Metrics}
This study employs five core metrics to comprehensively evaluate model performance: Attack Presentation Classification Error Rate (APCER) measures security; Bona Fide Presentation Classification Error Rate (BPCER) reflects usability; Average Classification Error Rate (ACER) characterizes the overall discriminative level at a specific threshold; Area Under the ROC Curve (AUC) depicts the global ranking capability across thresholds. Given the inherent trade-off between APCER and BPCER, multiple metrics must be combined to comprehensively assess the balancing capability between security and usability\cite{yu2023face}.
\subsubsection{Evaluation Dataset}
To thoroughly evaluate the ID generalization ability and OOD transfer robustness, the test set consists of four independent components:
\begin{itemize}
    \item \textbf{CelebA-Spoof (ID):} CelebA-Spoof is a large-scale face anti-spoofing dataset comprising 625,537 images across 10,177 identities, covering diverse capture devices and scenarios. The spoof samples encompass 11 types of attack methods including print, 3D mask, and replay attacks, and each image is annotated with 43 fine-grained attributes, making it one of the widely adopted benchmarks in this field\cite{zhang2020celeba}.
    \item \textbf{MS-UFAD-DeepFake (ID):} MS-UFAD is a large-scale unified face attack detection dataset comprising 795k videos and 60k images, covering 5,000 identities and 52 attack methods across four quality levels. It provides fine-grained textual descriptions for each sample through semi-automated annotation, making it the first face attack dataset to incorporate textual semantic annotations. We adopt its digital forgery subset for model training and evaluation\cite{jiang2025ms}.
    \item \textbf{MFFI-Phase-Val (OOD):} MFFI is a multi-dimensional face forgery image dataset tailored for real-world scenarios, comprising 1024K image samples and 50 different forgery methods. It enhances realism across four strategic dimensions—wider forgery methods, varied facial scenes, diversified authentic data, and multi-level degradation operations—and outperforms existing datasets in scene complexity, cross-domain generalization, and detection difficulty gradients\cite{miao2025mffi}. We exclusively adopt its test set to evaluate the out-of-distribution (OOD) generalization capability of our model.
\end{itemize}
\subsubsection{Training Configurations}
The fine-tuning configuration strikes a balance between parameter efficiency, context adaptability, and convergence sufficiency. We employ a LoRA strategy, which injects low-rank update matrices into all linear layers. This approach significantly reduces memory overhead and the number of trainable parameters while preserving the model's representation capability for downstream tasks. The sequence length is set to 2048, providing a sufficient context window for the MLLMs to accommodate multimodal tokens, thereby effectively preventing information loss caused by truncation. Furthermore, combined with 3 training epochs and a 5\% warmup ratio, the configuration ensures the model adequately learns instruction-following patterns on the subset data while mitigating the risks of overfitting and catastrophic forgetting.

\subsection{Ablation Study}\label{sec:as}
To validate the effectiveness of CoT, we present two pivotal ablation studies. First, we compare our approach against large-scale short-text instruction fine-tuning in Section \ref{sec:cebcf}. Next, we evaluate the robustness of CoT fine-tuning under white-box attacks relative to short-text instruction fine-tuning in Section \ref{sec:cracs}. Furthermore, since our evaluation involves both human experts and automated metrics, we present a comparative analysis between human and large model evaluations of CoT in Section \ref{sec:her}.  Finally, we provide the tuning results for the LoRA rank, which is a core hyperparameter in the SFT phase in Section \ref{sec:lce}. 

In the white-box attack experiments, we observe a pronounced trade-off between robustness and generalization, and we attempt to mitigate this issue via reinforcement learning. The generalization capability and adversarial robustness of face anti-spoofing models are of great practical significance, and addressing this challenge constitutes our primary future work.
\subsubsection{Comparative Experiments between CoT Fine-tuning and Large-scale Short-text Instruction Fine-tuning}\label{sec:cebcf}
\begin{figure}[htbp]
    \centering
    
    \begin{subfigure}[b]{0.3\linewidth}
        \centering
        \includegraphics[width=\linewidth]{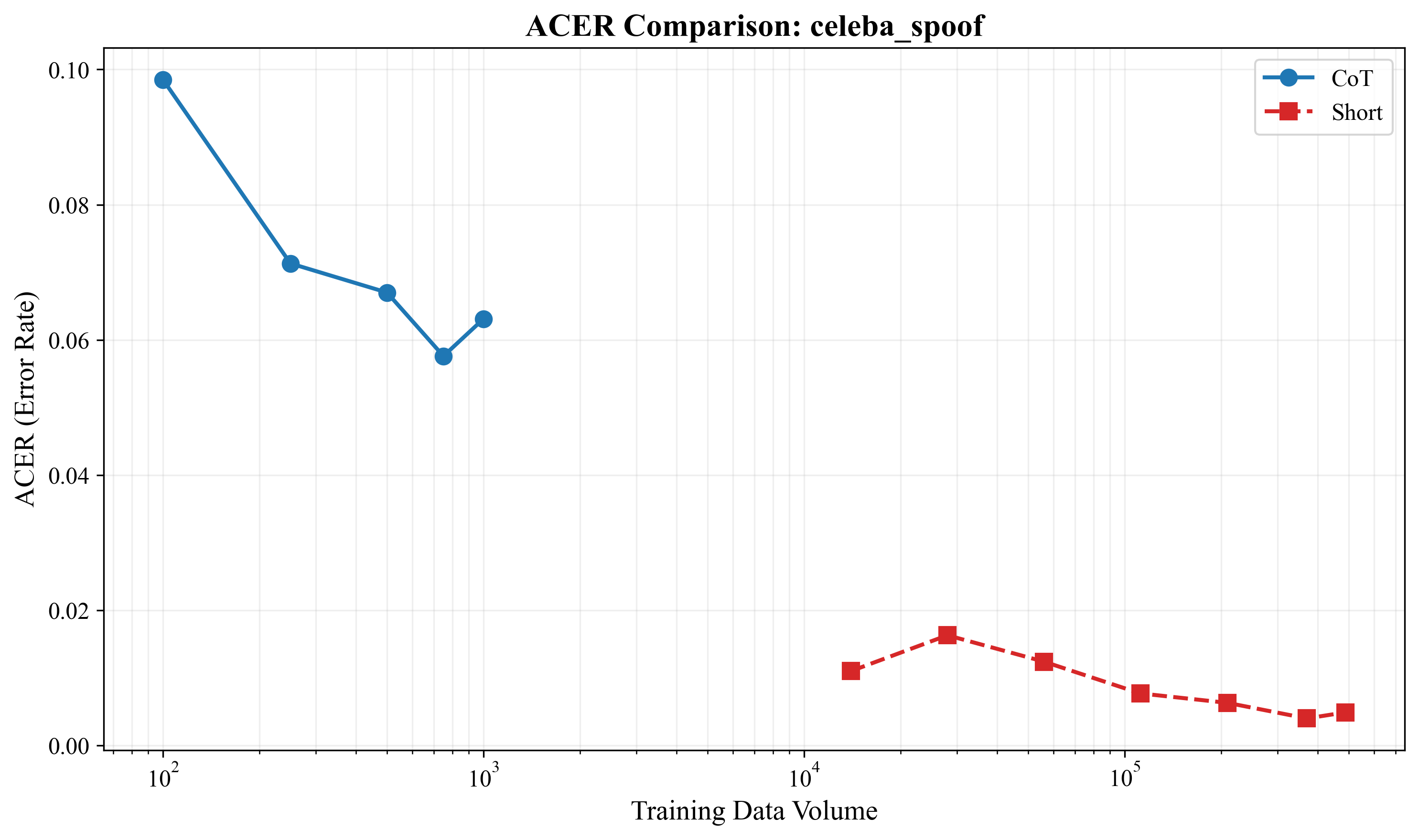}
        \caption{ACER trend comparison chart on the Celeba-Spoof dataset.}
        \label{fig:sub1}
    \end{subfigure}
    \hfill 
    \begin{subfigure}[b]{0.3\linewidth}
        \centering
        \includegraphics[width=\linewidth]{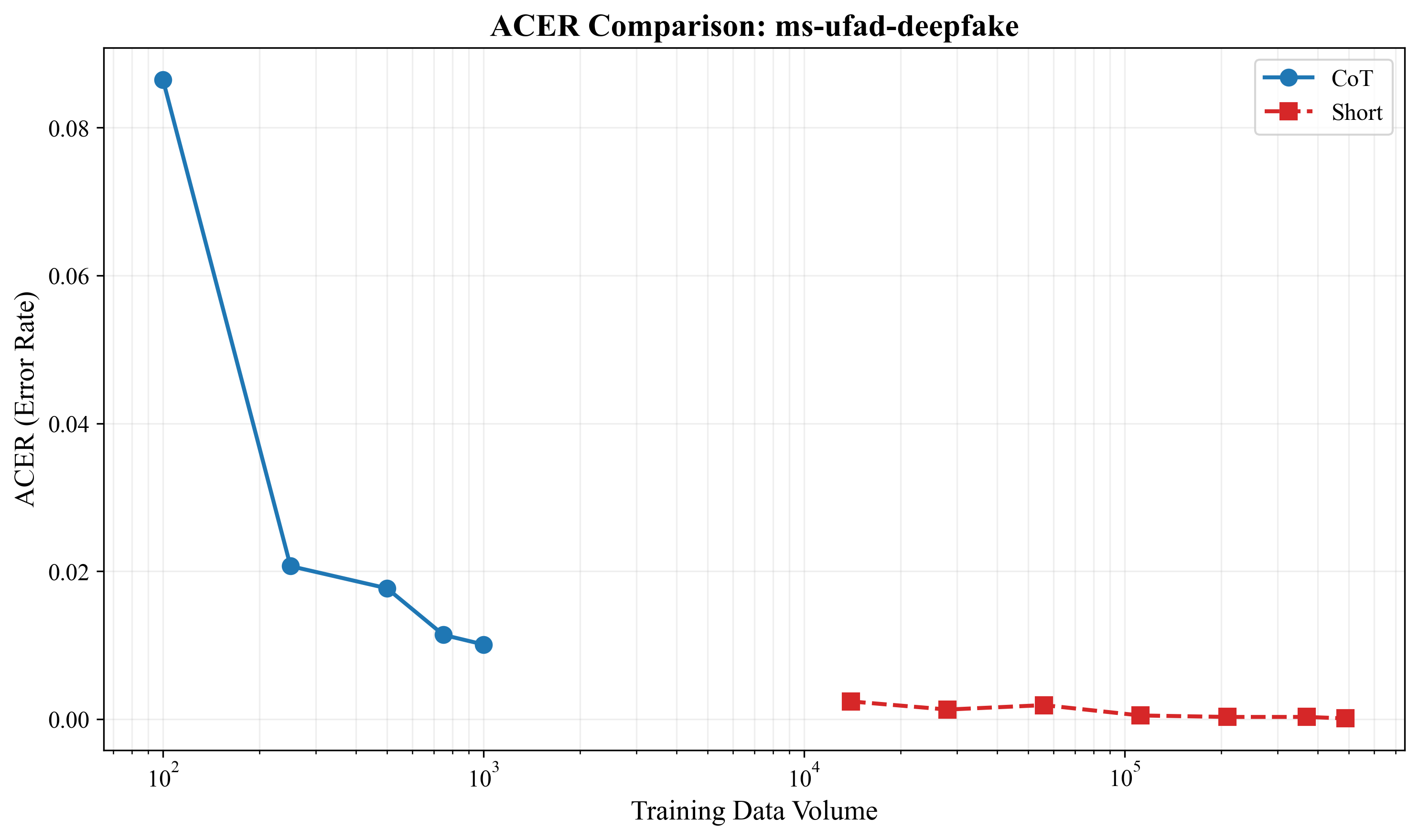}
        \caption{ACER trend comparison chart on the MS-UFAD-Deepfake dataset.}
        \label{fig:sub2}
    \end{subfigure}
    \hfill
    \begin{subfigure}[b]{0.3\linewidth}
        \centering
        \includegraphics[width=\linewidth]{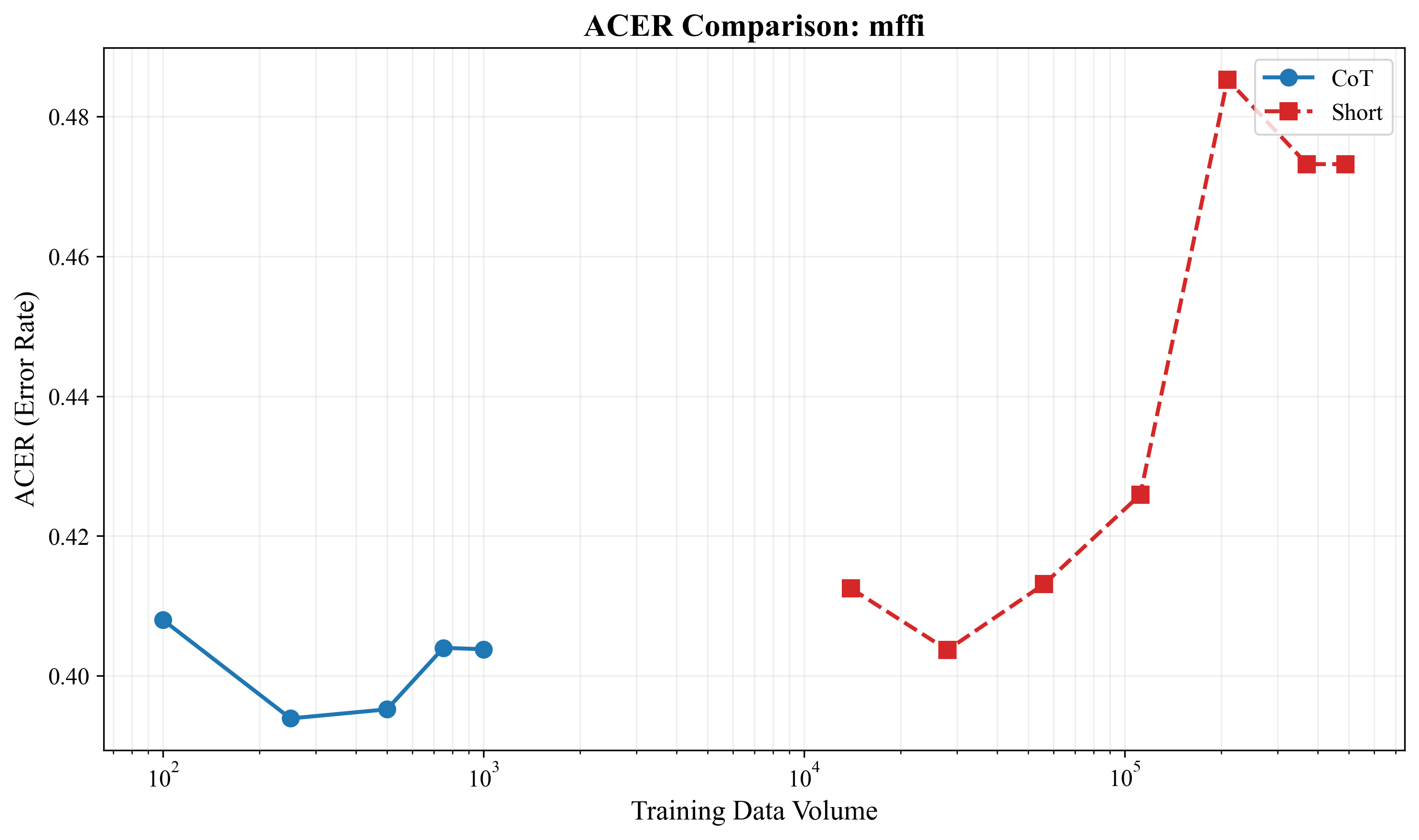}
        \caption{ACER trend comparison chart on the MFFI dataset.}
        \label{fig:sub3}
    \end{subfigure}
    
    \vspace{0.5cm} 
    
    \begin{subfigure}[b]{0.3\linewidth}
        \centering
        \includegraphics[width=\linewidth]{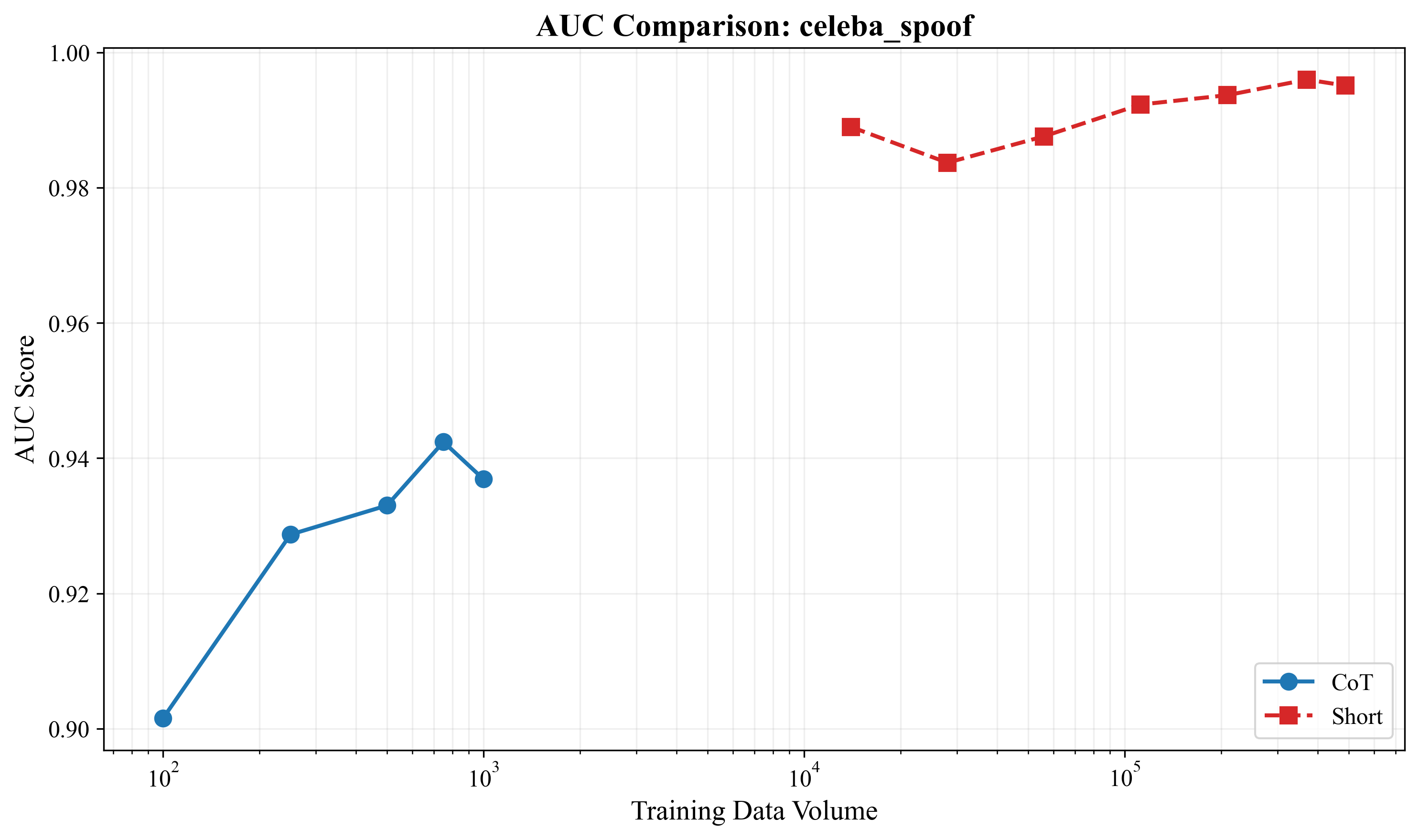}
        \caption{AUC trend comparison chart on the Celeba-Spoof dataset.}
        \label{fig:sub4}
    \end{subfigure}
    \hfill
    \begin{subfigure}[b]{0.3\linewidth}
        \centering
        \includegraphics[width=\linewidth]{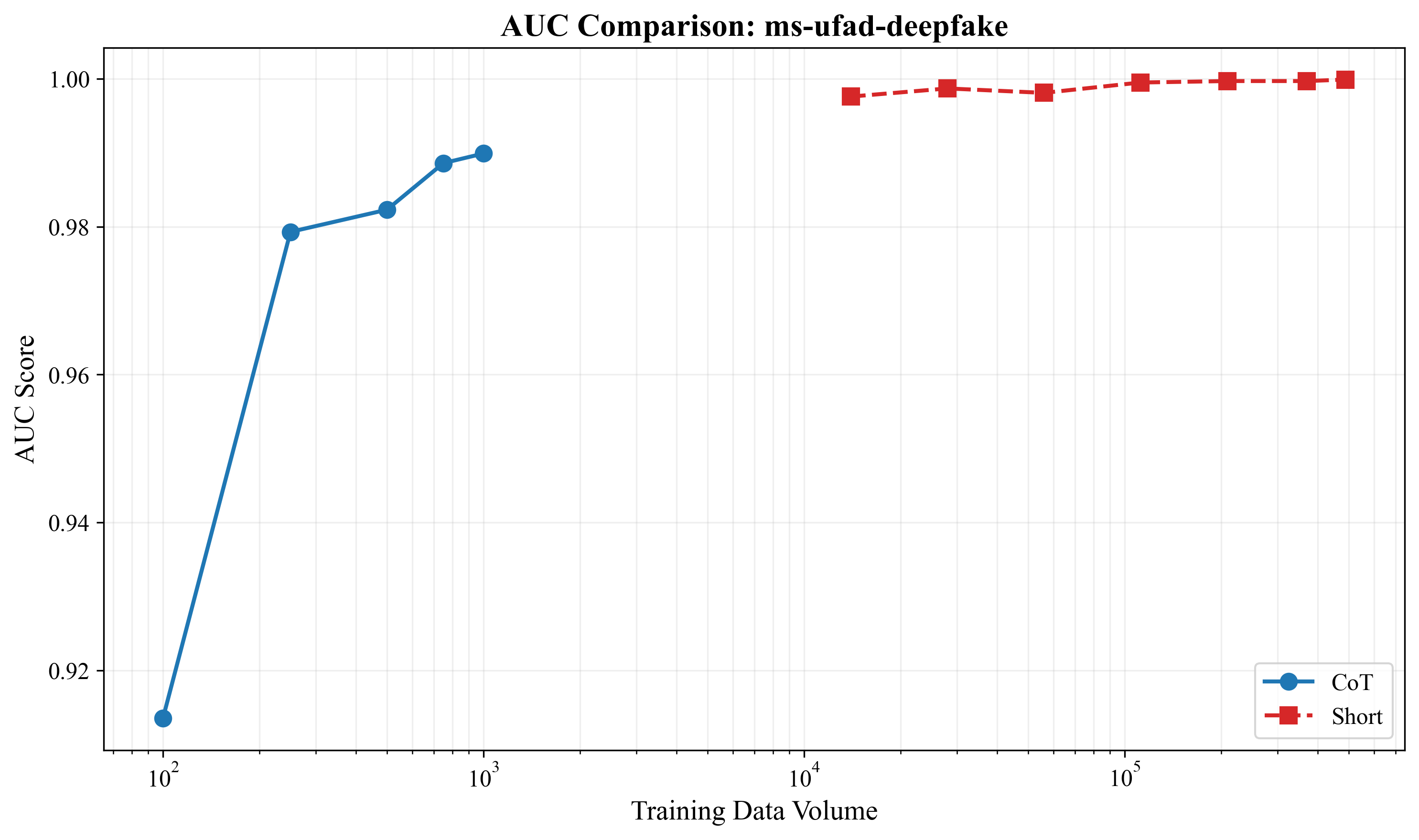}
        \caption{AUC trend comparison chart on the MS-UFAD-Deepfake dataset.}
        \label{fig:sub5}
    \end{subfigure}
    \hfill
    \begin{subfigure}[b]{0.3\linewidth}
        \centering
        \includegraphics[width=\linewidth]{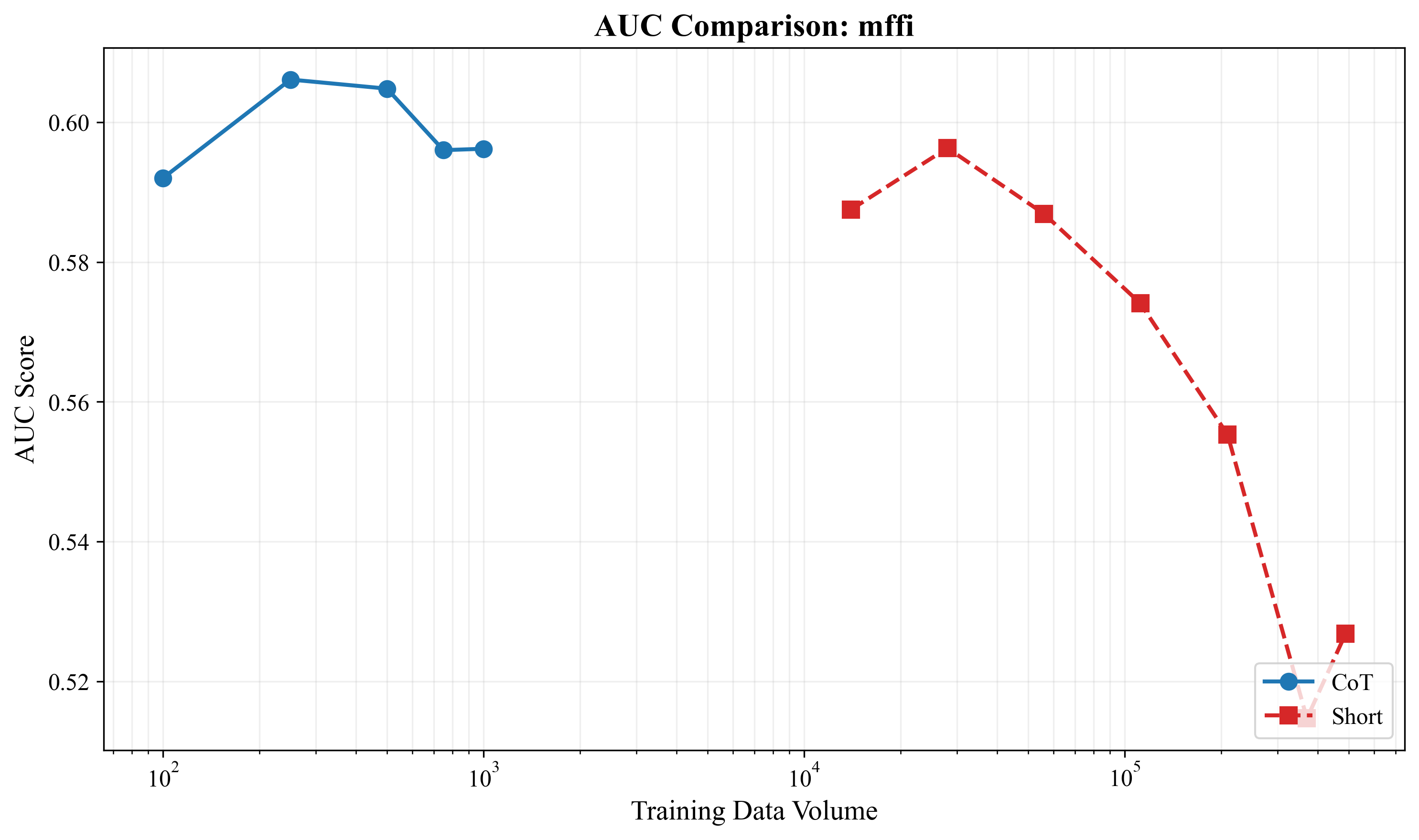}
        \caption{AUC trend comparison chart on the MFFI dataset.}
        \label{fig:sub6}
    \end{subfigure}
    
    \caption{Comparative experiments on model performance as a function of training data scale for CoT and short-text datasets.}
    \label{fig:all_six}
\end{figure}

To validate the effectiveness of our CoT approach, we first conduct a comparative experiment against short-text fine-tuning. Notably, short-text fine-tuning can be regarded as an approximate upper bound for methods such as MS-UFAD\cite{jiang2025ms}\cite{fang2024unified}\cite{zheng2026unified}. The experimental results are illustrated in Figure \ref{fig:all_six}. On in-domain evaluation, CoT data achieves comparable performance to fine-tuning on hundreds of thousands of short-text samples using only a few thousand samples. On the CelebA-Spoof and MS-UFAD-Deepfake datasets, as the scale of short-text training data expands, the model achieves competitive performance that slightly surpasses the CoT fine-tuning baseline. However, this requires a data volume dozens of times larger than that of the CoT approach. This suggests that the short-text model can effectively extract shallow statistical regularities and specific feature shortcuts within the training set. Meanwhile, the CoT model also maintains strong performance on ID data (with AUCs ranging from 0.94 to 0.99). This demonstrates that the introduction of logical constraints by the reasoning mechanism during feature extraction does not significantly compromise absolute metrics. More importantly, it prevents the model from excessively memorizing noise in the training set.

However, when the evaluation shifts to the OOD MFFI dataset, the generalization capabilities of the two architectures exhibit significant heterogeneity. The AUC of the model trained on short texts on MFFI plummets to the range of 0.51 to 0.59, remaining strictly below 0.6 and indicating near-random guessing. In stark contrast, the CoT model achieves an AUC of 0.67, significantly surpassing the short-text fine-tuned data and demonstrating a certain degree of discriminative capability. The near-random-guessing performance of the former confirms its heavy reliance on the specific distribution of the training data. Once confronted with distribution shifts, its memorized feature shortcuts become completely ineffective and may even trigger negative transfer. Conversely, the superior performance of the CoT model demonstrates that the logical reasoning mechanism introduced via CoT prompts the model to discard superficial pixel-level features and instead learn underlying semantic logic with greater universality and resistance to interference, thereby preserving basic discriminative capabilities in unknown distributions.

Synthesizing these experimental results, we can conclude that the current learning paradigm of models is undergoing a profound shift from memorization to reasoning. The short model represents traditional intuitive pattern matching, which lacks the robustness to handle the complexities of the open world. The CoT model successfully trades off some ultimate fitting precision on ID tasks for generalization capabilities in OOD scenarios. This finding provides crucial insights for future development of highly robust systems: in real-world applications facing potential distribution shifts, introducing reasoning mechanisms to break the bottleneck of memorization is a key pathway to enhancing a model's survival and adaptability in unknown environments.

\subsubsection{Comparative Robustness Analysis of CoT and Short Strategies against Adversarial Physical and Digital Attacks}\label{sec:cracs}
To evaluate the robustness of the model, we employ a targeted white-box attack on the pixel space\cite{madry2018towards}. Given a face image $\mathbf{x}$, the adversarial image $\mathbf{x}^*$ is generated by iteratively maximizing the probability of a target token $y_{target}$ (e.g., class `real'). At each iteration step $t$, the perturbation is updated using the sign of the gradient of the cross-entropy loss $\mathcal{L}$ with respect to the input pixels:
\begin{equation}
    \mathbf{x}_{t+1} = \Pi_{\mathbf{x}, \epsilon} \left( \mathbf{x}_{t} + \alpha \cdot \text{sgn} \left( \nabla_{\mathbf{x}} \mathcal{L}(\mathbf{x}_{t}, y_{target}) \right) \right)
\end{equation}
where $\alpha$ is the step size, $\epsilon$ is the maximum allowable perturbation bound, and $\Pi_{\mathbf{x}, \epsilon}(\cdot)$ denotes the projection operation that clamps the perturbed pixels to the valid range $[\mathbf{x} - \epsilon, \mathbf{x} + \epsilon]$. The final adversarial image $\mathbf{x}^*$ is obtained after $T$ iterations.

\begin{figure}[htbp]
    \centering
    \includegraphics[width=0.85\linewidth]{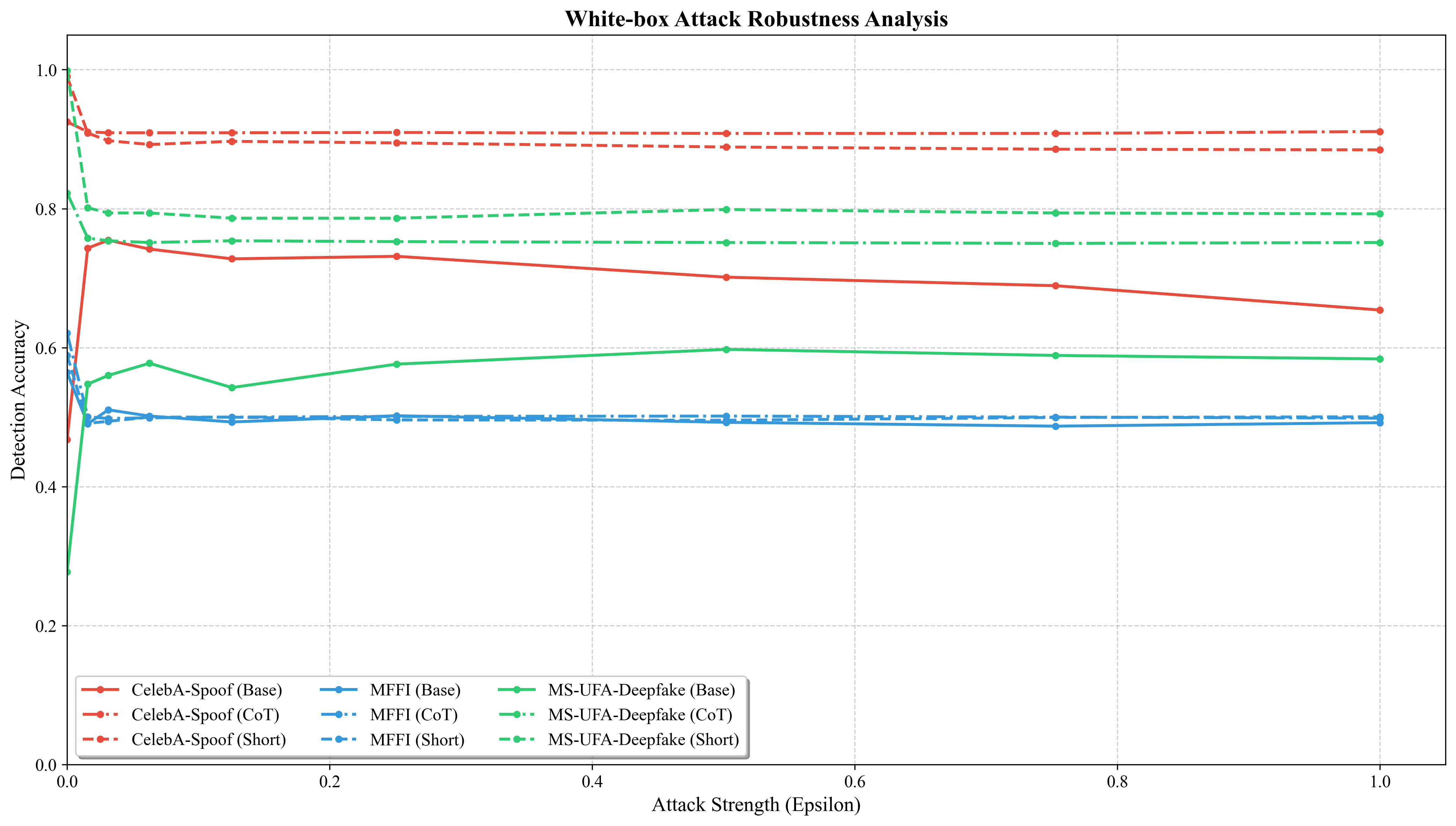}
    \caption{Comparison of the base model, the model fine-tuned on a large corpus of short-text samples, and the model fine-tuned on a small amount of CoT data under white-box attacks.}
    \label{fig:whitebox_attack_curves_fixed}
\end{figure}

The experimental results are shown in Figure \ref{fig:whitebox_attack_curves_fixed}. On the CelebA-Spoof dataset, which focuses on physical spoofing (e.g., prints and screen replays), the short-text fine-tuning strategy exhibits severe robustness vulnerabilities, whereas CoT demonstrates exceptional stability. While the short-text fine-tuning model achieves an extremely high accuracy of $99.00\%$ under no-attack conditions compared to $92.50\%$ for CoT, this suggests that the short-text fine-tuning model may be overfitting to specific physical textures or shallow features. As the intensity of white-box attacks increases, the accuracy of the short-text fine-tuning model plummets from $99.00\%$ to $88.45\%$, representing an absolute drop of $10.55\%$. In contrast, the accuracy of the CoT model only marginally decreases from $92.50\%$ to $91.09\%$, with an absolute drop of merely $1.41\%$. Under physical attacks, adversarial samples typically disrupt physical spoofing traces (such as moiré patterns and edge artifacts) through subtle pixel perturbations. Due to its excessively short-text fine-tuning reasoning path, the short-text fine-tuning strategy is easily blinded by such perturbations. Conversely, CoT constructs a redundant verification mechanism through multi-step reasoning (e.g., global lighting analysis and biometric consistency checks), successfully defending against physical gradient attacks.

In the MS-UFAD-Deepfake dataset, which targets digitally generated forgeries, the defensive advantage of CoT is further amplified. The initial accuracy of the short-text fine-tuning model is as high as $99.87\%$, but it instantly drops to $80.12\%$ under a mild attack ($\varepsilon=0.0157$) and continues to decline to $79.25\%$ in subsequent attacks, with a maximum drop exceeding $20\%$. This indicates that short logical chains are highly fragile when confronted with high-frequency synthetic artifacts in the digital domain. The initial accuracy of CoT is $82.25\%$. After an initial adaptive adjustment, its accuracy remains strictly locked within the $75.00\%$--$75.37\%$ range throughout the entire high-intensity attack interval ($\varepsilon=0.0627$ to $\varepsilon=1.0000$). Since digital attacks typically inject high-frequency noise directly into the feature space, the long-chain reasoning of CoT acts as a semantic low-pass filter. This mechanism filters out digital adversarial perturbations, maintaining a highly robust performance floor.

On the cross-domain MFFI dataset, the accuracy of Base, CoT, and Short models all rapidly plummet to approximately $50\%$, which is the level of random guessing. This result inversely validates the boundary of effectiveness: the robustness gains of CoT are highly dependent on the model's existing cognitive representations of target spoofing features. When confronted with completely unknown OOD data, the lack of correct prior knowledge as a reasoning foundation causes multi-step reasoning to lose its factual basis, thereby rendering the defense mechanism ineffective.

Overall, under both physical and digital attacks, the accuracy degradation of CoT ($1.41\%$ and $7.12\%$) is substantially lower than that of the short-text fine-tuning strategy ($10.55\%$ and $20.62\%$). CoT not only defends against high-frequency adversarial noise in the digital domain but also effectively resists texture-level adversarial perturbations in the physical domain, demonstrating its immense potential as a universal semantic-level defense mechanism.

\subsubsection{Human Evaluation Results}\label{sec:her}
This experiment establishes an automated quality assessment protocol based on the GPT-4.1 MLLM, aiming to audit CoT annotations in FAS datasets. To prevent misjudgments on genuine live samples that lack obvious forgery traces, the experiment introduces a dynamic baseline calibration mechanism. When an image is identified as genuine and both logical and knowledge scores are passing, the system forcibly sets a floor of 3 for the D1 score. In terms of data presentation, the experiment ultimately generates a structured quality sampling report in a tabular format, recording detailed information for each sample in rows, including the filename, D1 evidence score, D2 logical score, D3 knowledge score, the comprehensive score calculated via a weighted formula, the final pass/fail verdict, and a summary of critical issues output by the model. Through this evaluation framework, the experiment can accurately identify specific flaws such as incorrect conclusions, hallucinated features, or omission of secondary clues, thereby providing quantitative evidence for the refined iteration of the dataset.

We sampled 100 instances from each of the final generated CoT categories and each attack type for model evaluation. The experimental results are presented in Table \ref{tab:evaluation_scores}. The human expert CoT evaluation achieves the highest weighted score of 4.7619, followed by GPT-4.1 with 4.7046, which outperforms the human expert on D2 (4.9043) and D3 (4.7186). Qwen-3.6 obtains the lowest scores across all metrics, with a weighted score of only 4.2536. Even when a CoT reasoning chain contains minor flaws, human experts are able to recognize the overall soundness of the reasoning, whereas models tend to perform strict step-by-step verification and are prone to penalizing the overall score due to local errors. During evaluation, large language models may place greater emphasis on surface-level features such as structural completeness and logical coherence of the CoT, resulting in harsher scoring for reasoning chains that deviate from their ideal templates. Human experts can tolerate certain non-standard yet effective reasoning paths based on their experience, while models lack such flexible judgment capability.

\begin{table}[htbp]
  \centering
  \caption{Comparison of Evaluation Scores across Different Models and Human Experts}
  \label{tab:evaluation_scores}
  \renewcommand{\arraystretch}{1.3} 
  \begin{tabular}{lcccc}
    \toprule
    \textbf{Evaluator} & \textbf{D1 Avg.} & \textbf{D2 Avg.} & \textbf{D3 Avg.} & \textbf{Weighted Score} \\
    \midrule
    Human Expert       & 4.8024           & 4.7917           & 4.6476           & 4.7619                  \\
    GPT-4.1            & 4.5607           & 4.9043           & 4.7186           & 4.7046                  \\
    Qwen-3.6           & 4.0822           & 4.3987           & 4.3621           & 4.2536                  \\
    \bottomrule
  \end{tabular}
\end{table}

\subsubsection{LoRA Configuration Experiments}\label{sec:lce}
\begin{figure}[htbp]
    \centering
    \includegraphics[width=0.85\linewidth]{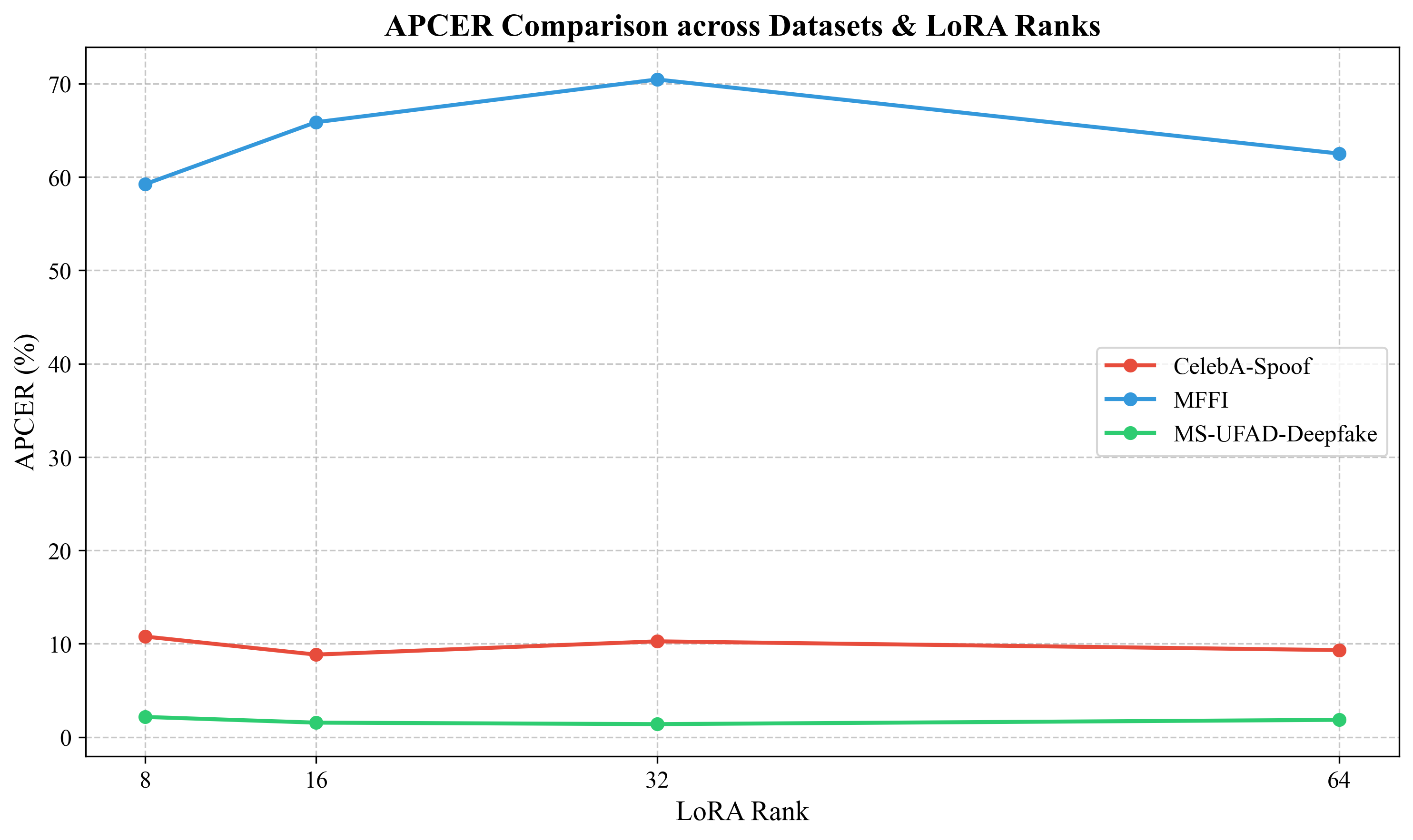}
    \caption{Performance analysis of LoRA configurations across physical, digital, and cross-domain anti-spoofing tasks.}
    \label{fig:lora_experiment}
\end{figure}

Figure \ref{fig:lora_experiment} illustrates the impact of different LoRA rank settings on the APCER across various datasets. On the CelebA-Spoof and MS-UFAD-Deepfake datasets, the APCER remains consistently low and exhibits minimal fluctuation across varying LoRA ranks. In contrast, the APCER on the OOD dataset, MFFI, is significantly higher than that of the ID data, fluctuating between 60\% and 70\%. Notably, as the rank increases from 8 to 32, the APCER rises, peaking at a rank of 32. This suggests that with increased LoRA capacity, the model may overfit the training set, learning features that are overly specific to the source domain, thereby compromising its generalization capability on MFFI. However, as the rank further increases to 64, the APCER experiences a slight decline. Under this larger capacity, the model may have sufficient space to learn more generalized or robust features, or this improvement might simply be attributed to a saturation effect of overfitting.

For OOD generalization, a rank of 8 appears to be a relatively optimal choice, despite the high absolute error rate. This corroborates the hypothesis that under severe domain shifts, restricting model capacity can help prevent overfitting to source domain noise, thereby preserving a degree of generalizability. While LoRA fine-tuning is highly effective on the source domain, easily achieving low error rates, OOD performance proves to be highly sensitive to the LoRA rank, unlike the stability observed in ID tasks. Ultimately, our empirical results demonstrate that blindly increasing the LoRA rank does not enhance cross-domain generalization. Instead, it may further degrade generalization performance due to overfitting to source domain features. In this specific cross-domain setting, a smaller rank demonstrates relatively superior robustness.

\subsection{Comparative Experiments }\label{sec:ce}
\subsubsection{Comparative Experimental Setup}
The field of face anti-spoofing is currently in a transitional phase where CNNs, CLIP, and MLLMs coexist. Many state-of-the-art (SOTA) methods are built upon traditional CNNs or foundational vision-language models (e.g., CLIP). However, due to the proprietary nature of their training data and undisclosed source codes, exact reproduction of these specific variants is unfeasible. Furthermore, these incremental improvements typically do not break through the inherent performance ceilings of their respective backbone architectures. Therefore, rather than relying on unreproducible specific implementations, we employ ResNet-101 and CLIP-ViT-B as the representative upper bounds for their respective methodological categories. This strategy ensures that our comparative evaluation remains both rigorous and reproducible. In our comparative experiments, we evaluate four categories of baseline models. In terms of general capability, they are ranked from strongest to weakest as follows:

\begin{enumerate}
    \item \textbf{Traditional CNNs}: Represented by ResNet-50 and ResNet-101. These methods serve as a representative baseline for conventional deep learning approaches\cite{kuznetsov2024attacknet}\cite{jingade2023extended}\cite{liu2018learning}\cite{rossler2019faceforensics++}\cite{lin2019face}\cite{deb2023unified}.
    \item \textbf{Foundation Multimodal Models}: Specifically, CLIP-ViT-B/32. This model represents the general scope of prior multimodal methods\cite{narayan2025facexformer}\cite{lin2025reliable}\cite{sun2025towards}.
    \item \textbf{Open-source Large Models}: Including Qwen3-235B-A22B-Thinking-2507, Qwen3-VL-32B-Instruct, Qwen3-VL-8B-Thinking, Qwen3-VL-3B-Instruct, and Qwen3-VL-7B-Instruct\cite{Qwen-VL}.
    \item \textbf{Proprietary Large Models}: Including GPT-4o, Gemini-2.5-Flash-Image, and Claude-Opus-4.6.
\end{enumerate}

Next, we first analyze the necessity of employing multimodal large models, and subsequently conduct a comprehensive comparison with various baseline models. The necessity analysis primarily involves a detailed comparison and comprehensive analysis against traditional CNNs.
\subsubsection{Necessity of Adopting MLLMs for the UFAD Task}
\begin{figure}[htbp]
    \centering
    \includegraphics[width=0.85\linewidth]{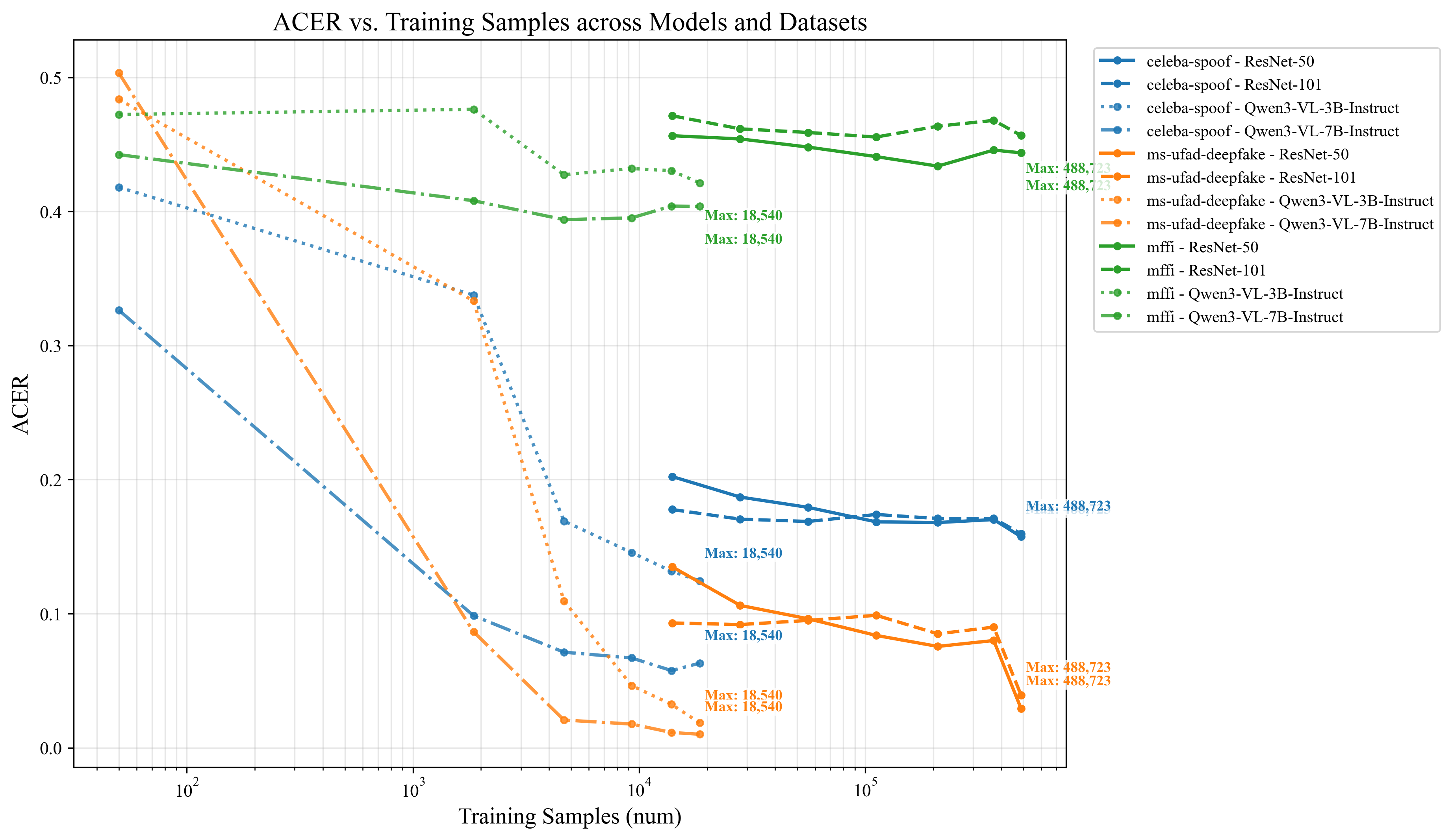}
    \caption{Performance analysis of LoRA configurations across physical, digital, and cross-domain anti-spoofing tasks.}
    \label{fig:cnn_compare_experiment}
\end{figure}

Compared to traditional CNNs, which rely solely on shallow feature matching of low-level pixels and frequency-domain artifacts, MLLMs demonstrate a generational advantage in joint face anti-spoofing. MLLMs overcome the detection bottlenecks of unimodal models under compression and noise interference. By leveraging cross-modal attention mechanisms, they precisely capture subtle logical contradictions in audio-visual semantics. Furthermore, empowered by massive pre-trained knowledge, they achieve superior generalization against unseen spoofing methods. Most importantly, MLLMs abandon the black-box paradigm of traditional methods, providing interpretable forensic evidence in natural language, thereby realizing a true paradigm shift from mere texture recognition'' to cognitive understanding.  To this end, we conduct a comparative evaluation against CNN methods across varying orders of magnitude of training samples. The experimental results are illustrated in Figure \ref{fig:cnn_compare_experiment}.

As shown in Figure \ref{fig:cnn_compare_experiment}, traditional CNN methods, represented by ResNet-50 and ResNet-101, exhibit significant limitations in face anti-spoofing tasks. Although they achieve remarkably low error rates on specific datasets (with an ACER as low as $0.0756$), their performance degrades drastically in cross-domain scenarios (e.g., MFFI), where the ACER soars to over $0.43$. This indicates a severe tendency toward overfitting and inadequate generalization capabilities. In contrast, the MFAD series of specialized lightweight anti-spoofing models (e.g., MFAD-7B), through targeted architectural design, not only maintains an exceptionally low error rate in-domain (ACER of $0.0357$ on CelebA-Spoof) but also significantly outperforms generic ResNet models in cross-domain testing. This demonstrates that dedicated architectural designs tailored for anti-spoofing tasks are substantially more effective than merely increasing the depth of CNNs.

\subsubsection{Comparison with All Model Types}
As a general-purpose vision-language pre-trained model, the CLIP-ViT-base series exhibits a distinct characteristic of strong in-domain performance but weak cross-domain generalization in face anti-spoofing tasks. On the MS-UFAD-Deepfake dataset, CLIP demonstrates remarkable artifact-capturing capabilities (with the patch32 variant achieving an ACER of merely $0.0084$), even outperforming certain large-scale multimodal models. However, it yields mediocre results on CelebA-Spoof, and its performance degrades drastically in the cross-domain MFFI test, where the ACER soars to approximately $0.48$, approaching the level of random guessing. This indicates that although CLIP has learned rich semantic representations, it lacks sensitivity to fine-grained anti-spoofing textures. Furthermore, the inherent gap between its general pre-training objectives and the specific task of spoof detection leads to significantly inferior generalization capabilities against unseen domains compared to specialized models.

On the ID datasets, general-purpose models without targeted training exhibit clear limitations, whereas the specifically trained MFSA model demonstrates a superior performance ceiling. Despite possessing massive parameter counts, general large models such as Qwen3-VL-32B-Instruct and Gemini-2.5-Flash-Image generally yield ACERs higher than 0.15, reaching up to 0.36 on CelebA-Spoof and MS-UFAD-Deepfake. In contrast, MFSA-7B reduces the ACER to 0.0357 and 0.0101, respectively. This proves that relying solely on the pre-trained knowledge of general models is insufficient to effectively capture the subtle artifacts of deepfakes. Thus, targeted supervised training is essential to inject domain-specific knowledge. Although ResNet-50/101 achieve low APCER on MS-UFAD, they suffer from high BPCER (approximately 0.19), indicating a difficulty in balancing the classification of genuine and spoof samples. Meanwhile, MFAD-7B maintains extremely low APCER (0.0715 / 0.0167) while achieving 0 BPCER on CelebA-Spoof and 0.0035 BPCER on MS-UFAD. This indicates that our training strategy not only improves detection rates but also significantly optimizes the decision boundary, thereby avoiding false rejections of genuine samples.

The experimental results clearly illustrate the performance gains of the MFAD series models with increasing parameter scales, validating the effectiveness of our proposed method. From MFAD-3B to MFAD-7B, all metrics show significant improvement. Specifically, on MS-UFAD-Deepfake, the ACER is further compressed from 0.0186 to 0.0101. This consistent performance enhancement suggests that our training data and loss function design are of high quality and can be effectively leveraged by larger-scale models, providing a solid empirical foundation for future model scaling. Although MFAD-7B exhibits a higher ACER (0.359) on the out-of-domain dataset MFFI compared to some general models, this precisely reflects the fundamental difference in design objectives between specialized and general models, rather than a defect. To achieve an extremely low false rejection rate (BPCER $\approx$ 0) on in-domain data, the MFAD model learns a highly fine-grained ID feature distribution. This high specificity is the cause of its performance degradation on the unseen MFFI domain. In practical application scenarios (such as financial identity verification and social media moderation), the threats faced are typically specific types of attacks. In such cases, clients prioritize zero false rejections and high detection rates for known attacks. The overwhelming advantage demonstrated by the MFAD model on ID data is exactly what is urgently needed for real-world deployment. While general models may perform adequately on OOD data, their error rates of 20--30\% on ID data are unacceptable in industrial applications.

\begin{table}[htbp]
  \centering
  \caption{Performance comparison of different models across CelebA-Spoof, MS-UFAD-Deepfake, and MFFI datasets.}
  \label{tab:model_comparison}
  
  \tiny 
  \setlength{\tabcolsep}{1.5pt}
  \renewcommand{\arraystretch}{1.1} 
  
  \begin{tabular*}{\textwidth}{@{\extracolsep{\fill}}lccccccccc}
    \toprule
    
    & \multicolumn{3}{c}{\textbf{CelebA-Spoof}} 
    & \multicolumn{3}{c}{\textbf{MS-UFAD-Deepfake}} 
    & \multicolumn{3}{c}{\textbf{MFFI}} \\
    
    \cmidrule(r){2-4} \cmidrule(lr){5-7} \cmidrule(l){8-10}
    
    \textbf{Model} & \textbf{APCER} & \textbf{BPCER} & \textbf{ACER} 
                   & \textbf{APCER} & \textbf{BPCER} & \textbf{ACER} 
                   & \textbf{APCER} & \textbf{BPCER} & \textbf{ACER} \\
    \midrule
    ResNet-50 & 0.1885 & 0.1475 & 0.1680 & 0.0040 & 0.1475 & 0.0756 & 0.2382 & 0.6293 & 0.4338 \\
    ResNet-101 & 0.1505 & 0.1975 & 0.1740 & 0.0000 & 0.1975 & 0.0988 & 0.1735 & 0.7375 & 0.4555 \\
    CLIP-ViT-base-patch16	& 0.1191	& 0.1501	& 0.1346	& 0.0247	& 0.0081	& 0.0177	& 0.4764	& 0.5139	& 0.4951\\
    CLIP-ViT-base-patch32	& 0.1001	& 0.3055	& 0.2028	& 0.0107	& 0.0061	& 0.0084	& 0.633	& 0.3242	& 0.4786\\
    Qwen3-VL-8B-Thinking & 0.8032 & 0.0175 & 0.4104 & 0.7206 & 0.0175 & 0.3690 & 0.7147 & 0.0309 & 0.3728 \\
    Qwen3-VL-32B-Instruct & 0.4482 & 0.0425 & 0.2454 & 0.3288 & 0.0425 & 0.1856 & 0.5529 & 0.0385 & 0.2958 \\
    Qwen3-235B-A22B-Thinking-2507 & 0.7207 & 0.5263 & 0.6235 & 0.8575 & 0.5263 & 0.6919 & 0.4701 & 0.5676 & 0.5193 \\
    Qwen3-VL-3B-Instruct & 0.4029 & 0.4369 & 0.4199 & 0.5427 & 0.4375 & 0.4901 & 0.6756 & 0.2529 & 0.4643 \\
    Qwen3-VL-7B-Instruct & 0.5627 & 0.0673 & 0.3150 & 0.9127 & 0.0244 & 0.4686 & 0.7448 & 0.0778 & 0.4113 \\
    GPT-4o & 0.0701 & 0.2519 & 0.1610 & 0.0569 & 0.2519 & 0.1544 & 0.0152 & 0.9571 & 0.4861 \\
    Gemini-2.5-Flash-Image & 0.5400 & 0.0000 & 0.2700 & 0.6233 & 0.0152 & 0.3192 & 0.1512 & 0.2604 & 0.2058 \\
    Claude-Opus-4.6 & 0.3045 & 0.0125 & 0.1585 & 0.8427 & 0.0125 & 0.4276 & 0.3441 & 0.1815 & 0.2628 \\
    MFAS-3B & 0.1541 & 0.0000 & 0.0770 & 0.0340 & 0.0031 & 0.0186 & 0.5540 & 0.1422 & 0.3481 \\
    MFAS-7B & 0.0715 & 0.0000 & 0.0357 & 0.0167 & 0.0035 & 0.0101 & 0.6176 & 0.1004 & 0.3590 \\
    
    \bottomrule
  \end{tabular*}
\end{table}

In conclusion, training the MFAD model is not only necessary but also highly efficient. It resolves the issue of general large models understanding common sense but lacking anti-spoofing expertise in forgery detection tasks, while simultaneously overcoming the shortcomings of traditional CNNs, which suffer from high false rejection rates despite decent detection capabilities. Despite a performance drop in extreme OOD scenarios, its exceptionally high accuracy and ultra-low false rejection rates in core business scenarios (in-domain) firmly establish its core value as a specialized solution for forgery detection.
\section{Conclusion and Future Work}
In this paper, we have presented MFAD, a unified and interpretable face anti-spoofing system designed to address the critical challenges of generalization and interpretability in real-world scenarios. By leveraging the rich semantic priors of large vision-language models, our framework moves beyond traditional binary classification. The core of our approach lies in the Semantic-Level Evidence Anchoring mechanism, which grounds the decision-making process in human-understandable visual cues. Through the generation of CoT reasoning, MFAD not only achieves robust performance across various unseen domains but also provides transparent and trustworthy justifications for its predictions. Extensive experiments on standard benchmarks have demonstrated that our method significantly outperforms state-of-the-art approaches, establishing a new paradigm for developing reliable and interpretable biometric security systems.

We can observe that, on ID data, both robustness and accuracy have already reached a level suitable for practical deployment. However, performance on out-of-domain data remains insufficient: (1) the accuracy of OOD generalization still falls considerably short of the requirements for real-world applications; (2) there exists a pronounced trade-off between robustness and generalization. Addressing these challenges constitutes the focus of our future work.
\bibliographystyle{unsrtnat}

\section{Supplementary Material}
\subsection{Evaluation Guidelines for Human Experts, GPT-4o, and Qwen}
You are a senior FAS forensic analyst. Your task is to evaluate the reasoning quality of the  CoT text relative to the original image. 

\noindent \textbf{Core Principle:} The image is the sole ground truth. FAS datasets consist of two categories of samples: live and spoof.

\noindent \textbf{Mandatory Evaluation Protocol}
\begin{enumerate}
    \item \textbf{Independent Visual Analysis:} Evaluate the image \textit{without} referring to the text. Determine whether it is live and spoof, and list the key visual features.
    \item \textbf{Fact-Checking:} Compare the CoT content against your visual analysis point by point.
    \item \textbf{Dimension Decoupling:} Dimensions D1 (Factual), D2 (Logical), and D3 (Knowledge) must be scored independently.
\end{enumerate}

\noindent \textbf{Evaluation Dimensions}
\begin{description}
    \item[D1: Evidence Anchoring Accuracy (Weight: 30\%)] \leavevmode
    \begin{itemize}
        \item \textbf{5 Points:} Precisely describes spoof artifacts or genuine features, with a completely correct conclusion.
        \item \textbf{4 Points:} Core qualitative assessment is correct, with 1--2 minor descriptions being vague.
        \item \textbf{3 Points:} Core qualitative assessment is correct, but minor clues are omitted or slight descriptive deviations exist.
        \item \textbf{1 Point:} Core conclusion is incorrect, or non-existent features are hallucinated.
    \end{itemize}

    \item[D2: Reasoning Logical Coherence (Weight: 30\%)] \leavevmode
    \begin{itemize}
        \item \textbf{5 Points:} The reasoning chain is complete and closed-loop; every deduction is supported by prior observations.
        \item \textbf{3 Points:} Conclusion is correct and the overall logic is sound, but local logical jumps exist.
        \item \item \textbf{1 Point:} Logic is chaotic, circular reasoning is used, or the conclusion contradicts the observations.
    \end{itemize}

    \item[D3: Domain Knowledge Accuracy (Weight: 25\%)] \leavevmode
    \begin{itemize}
        \item \textbf{5 Points:} Terminology is precise, and the judgment aligns with standard FAS literature.
        \item \textbf{3 Points:} Judgment direction is correct, but terminology is non-standard.
        \item \textbf{1 Point:} Pseudo-scientific concepts are fabricated.
    \end{itemize}
\end{description}

\subsection{Human Modification Guidelines}
\begin{enumerate}
    \item \textbf{Error Diagnosis} \\
    Before initiating modifications, human experts must first accurately pinpoint the root cause of any breakdown or deviation within the Chain-of-Thought (CoT). Errors are typically categorized into three types:
    \begin{itemize}
        \item \textbf{Fact Error:} The model observes incorrect visual features or hallucinates non-existent details (e.g., in FAS forensics, misinterpreting shadows in a genuine photo as mask edges).
        \item \textbf{Logic Error:} The model possesses the correct knowledge but exhibits causal inversion, reasoning leaps, or circular reasoning during the deduction process.
        \item \textbf{Knowledge Error:} The model lacks specific domain expertise, leading to the application of incorrect evaluation criteria or tools.
    \end{itemize}

    \item \textbf{Direct Rewriting and Editing} \\
    This is the most straightforward rectification strategy. Human annotators preserve the valid initial observation steps in the CoT and commence targeted revision from the point of failure:
    \begin{itemize}
        \item \textbf{Supplementing Intermediate Reasoning Steps:} When the model jumps directly from facial texture observation to a liveness verdict, human experts insert the missing reasoning bridge (e.g., from specular reflection patterns to material-level analysis), thereby eliminating inferential gaps.
        \item \textbf{Correcting Reasoning Paths:} When the model misattributes natural facial shading to spoofing artifacts, human experts redirect the reasoning to the correct analytical path and explicitly articulate the self-correction process.
    \end{itemize}

    \item \textbf{Counterfactual Correction and Self-Reflection} \\
    A high-quality CoT should not only demonstrate how the correct conclusion is reached, but also explicate why alternative hypotheses are invalidated:
    \begin{itemize}
        \item \textbf{Falsification Logic Injection:} Human annotators incorporate counterfactual phrasing such as: ``Although moir\'e patterns were initially detected in the cheek region, frequency-domain verification ruled out a screen-replay attack.'' This instructs the model to perform cross-validation when confronting ambiguous facial artifacts.
    \end{itemize}
\end{enumerate}

\subsection{Exploration in Reinforcement Learning}
Having achieved promising results through SFT, we further employed the Group Relative Policy Optimization (GRPO)\cite{shao2024deepseekmath} algorithm to explore strategies for enhancing the OOD generalization capabilities. The experimental results are presented in Table \ref{tab:face_antispoof_perf}. The experimental results demonstrate that SFT is the decisive step in activating the face anti-spoofing capabilities of the Qwen3-VL-3B-Instruct model. During the SFT phase, ACER on in-domain datasets (CelebA-Spoof and MS-UFAD-Spoof) achieved a precipitous drop to 0.077 and 0.0186, respectively. Significant improvements were also observed on the out-of-domain dataset (MFFI), with the ACER decreasing to 0.3481, proving that SFT successfully endowed the model with foundational anti-spoofing representation capabilities. However, upon introducing the GRPO reinforcement learning phase, model performance failed to continue improving with the increase in sample size (from 200 to 60,000). Instead, severe performance degradation and catastrophic forgetting occurred. On the ID test sets, the ACER instantly regressed to a high plateau of approximately 0.42 and stagnated; on the out-of-domain MFFI test set, performance similarly deteriorated and showed a continuous worsening trend as the RL steps increased. This phenomenon indicates that despite the evaluation metrics in the RL phase being perfectly aligned with the test set, the GRPO algorithm under the current configuration not only failed to provide any gains but also severely disrupted the decision boundaries established by SFT.

\begin{table}[htbp]
    \centering
    \caption{Reinforcement Learning Experiments on MFAD-3B Fine-tuned from Qwen3-VL-3B-Instruct via SFT}
    \label{tab:face_antispoof_perf}
    
    
    \footnotesize 
    
    \renewcommand{\arraystretch}{1.0} 
    
    \setlength{\tabcolsep}{2pt} 
    
    \resizebox{\textwidth}{!}{%
        \begin{tabular}{l l c c c c c c c c} 
            \toprule
            \textbf{Dataset} & \textbf{Metric} & \textbf{Base} & \textbf{SFT} & \textbf{200} & \textbf{500} & \textbf{1k} & \textbf{10k} & \textbf{20k} & \textbf{40k} \\
            \midrule
            
            \multirow{3}{*}{CelebA-Spoof} 
            & APCER & 0.4029 & \textbf{0.1541} & 0.6208 & 0.6234 & 0.6259 & 0.6229 & 0.6248 & 0.623 \\
            & BPCER & 0.4369 & \textbf{0.0000} & 0.2179 & 0.2189 & 0.2271 & 0.2170 & 0.2184 & 0.217 \\
            & ACER  & 0.4199 & \textbf{0.0770} & 0.4194 & 0.4211 & 0.4265 & 0.4199 & 0.4216 & 0.420 \\
            \cmidrule(l){2-10} 
            
            \multirow{3}{*}{MS-UFAD-Spoof} 
            & APCER & 0.5427 & \textbf{0.0340} & 0.7453 & 0.7434 & 0.7422 & 0.7467 & 0.7443 & 0.743 \\
            & BPCER & 0.4375 & \textbf{0.0031} & 0.2224 & 0.2212 & 0.2191 & 0.2206 & 0.2228 & 0.222 \\
            & ACER  & 0.4901 & \textbf{0.0186} & 0.4839 & 0.4823 & 0.4806 & 0.4836 & 0.4836 & 0.482 \\
            \cmidrule(l){2-10} 
            
            \multirow{3}{*}{MFFI} 
            & APCER & 0.6756 & 0.5540 & 0.6567 & 0.6527 & 0.6580 & 0.6585 & 0.6526 & 0.653 \\
            & BPCER & 0.2529 & \textbf{0.1422} & 0.2907 & 0.2900 & 0.2876 & 0.2895 & 0.2898 & 0.290 \\
            & ACER  & 0.4643 & \textbf{0.3481} & 0.4737 & 0.4714 & 0.4728 & 0.4740 & 0.4712 & 0.471 \\
            
            \bottomrule
        \end{tabular}%
    } 

\end{table}

To address the performance collapse and catastrophic forgetting caused by GRPO, while simultaneously enhancing the generalization capabilities, interventions are recommended from algorithmic, data, and training strategy perspectives. First, regarding algorithmic stability, token-level importance sampling of GRPO in long-sequence generation often leads to high variance accumulation. We suggest introducing the Group Sequence Policy Optimization (GSPO) algorithm\cite{chujie2025group}, which shifts importance sampling to the sequence level and incorporates length normalization to fundamentally mitigate gradient instability. Second, to improve OOD generalization, it is essential to enrich the exploration space. We recommend employing diverse data augmentation techniques (e.g., geometric transformations, photometric distortion, or noise injection) during the RL sampling phase. This exposes the model to a broader distribution of spoofing artifacts and environmental variations, preventing it from overfitting to specific in-domain features.Third, to prevent the model from generating homogeneous samples within a group—which leads to a zero-variance gradient vanishing trap—we must enforce sampling diversity. This can be achieved by increasing the sampling temperature (e.g., to 0.6--1.0) and relaxing Top-P thresholds.Finally, to ensure RL gradient updates do not destroy SFT knowledge, we recommend mixing a proportion of original SFT data as a regularization constraint during GRPO training, or adding a distribution alignment step between SFT and RL to ensure the model conducts reinforcement exploration within a stable feature space.
\end{document}